\pdfoutput=1
\documentclass[letterpaper,twocolumn,10pt]{article}
\usepackage{usenix2019}

\usepackage{amsmath,amssymb,amsfonts}
\usepackage{algorithmic}
\usepackage{graphicx}
\usepackage{textcomp}
\usepackage{xcolor}
\usepackage{upgreek}
\usepackage{multicol}
\usepackage{multirow}
\usepackage{pgfplots}
\usepackage{textcomp}
\usepackage{tabu}
\usepackage{booktabs}
\usepackage{tikz}
\usepackage{xspace}
\usepackage{pgfplotstable}
\usepackage{subfig}
\usepackage[numbers]{natbib}
\usetikzlibrary{patterns}
\pgfplotsset{compat=newest}
\usepgfplotslibrary{groupplots}
\usepackage[export]{adjustbox}
\def\BibTeX{{\rm B\kern-.05em{\sc i\kern-.025em b}\kern-.08em
    T\kern-.1667em\lower.7ex\hbox{E}\kern-.125emX}}

\usepackage{url}
\usepackage[T1]{fontenc}
\usepackage[utf8]{inputenc}
\usepackage{frcursive}

\usepackage[ruled, vlined]{algorithm2e}
\SetKwFunction{FmakeModel}{createAttackModel}
\SetKwFunction{Fpredict}{predictMembership}
\SetKwFunction{FtrainProxy}{trainProxy}
\SetKwFunction{FtrainShadow}{trainShadow}
\SetKwFunction{FcalibrateThreshold}{calibrateThreshold}
\SetKwFunction{Fsample}{sample}
\SetKwFunction{Fsort}{sort}
\SetKwFunction{Fsplit}{split}
\SetKwProg{Fn}{def}{:}{}
\DontPrintSemicolon

\usepackage{caption}
\newtheorem{obs}{Observation}

\newtheorem{theorem}{Theorem}

\usepackage{setspace}
\usepackage[compact]{titlesec}
\titleformat{\paragraph}[runin]{\normalfont\bfseries}{}{0pt}{}[.]

\usepackage[subtle]{savetrees}

\newcommand{\N}{\mathcal N}

\newcommand{\p}[1]{\text{Pr}\left[#1\right]}
\newcommand{\sigmoidf}{\text{\cursive\slshape s}}
\newcommand{\sigmoid}[1]{\sigmoidf\left(#1\right)}
\newcommand{\half}{\frac{1}{2}}
\newcommand{\argmax}{\textit{argmax}}
\newcommand{\softmax}{\textit{softmax}}

\newcommand{\bayesian}{\textsf{bayes-wb}\xspace}
\newcommand{\bb}{\textsf{shadow-bb}\xspace}
\newcommand{\naive}{\textsf{naive}\xspace}
\newcommand{\general}{\textsf{general-wb}\xspace}
\newcommand{\omniscient}{\textsf{omniscient}\xspace}
\newcommand{\distrib}{\mathcal{D}\xspace}

\definecolor{bblue}{HTML}{4F81BD}
\definecolor{grayEight}{gray}{0.8}
\definecolor{graySix}{gray}{0.6}
\definecolor{grayFour}{gray}{0.4}
\definecolor{grayTwo}{gray}{0.2}

\begin{document}

\title{Stolen Memories: Leveraging Model Memorization for \\Calibrated White-Box Membership Inference}

\author{
{\rm Klas Leino}\\
Carnegie Mellon University
\and
{\rm Matt Fredrikson}\\
Carnegie Mellon University
} 

\maketitle

\begin{abstract}
\emph{Membership inference} (MI) attacks exploit the fact that machine learning algorithms sometimes leak information about their training data through the learned model.
In this work, we study membership inference in the \emph{white-box} setting in order to exploit the internals of a model, which have not been effectively utilized by previous work.
Leveraging new insights about how overfitting occurs in deep neural networks, we show how a model's idiosyncratic use of features can provide \emph{evidence for membership} to white-box attackers---even when the model's black-box behavior appears to generalize well---and demonstrate that this attack outperforms prior black-box methods.
Taking the position that an effective attack should have the ability to provide \emph{confident} positive inferences, we find that previous attacks do not often provide a meaningful basis for confidently inferring membership, whereas our attack can be effectively calibrated for high precision.
Finally, we examine popular defenses against MI attacks, finding that
\emph{(1)} smaller generalization error is not sufficient to prevent attacks on real models, and 
\emph{(2)} while small-$\epsilon$-differential privacy reduces the attack's effectiveness, this often comes at a significant cost to the model's accuracy; and for larger $\epsilon$ that are sometimes used in practice (e.g., $\epsilon=16$~\cite{applePrivacy}), the attack can achieve nearly the same accuracy as on the unprotected model.

\end{abstract}


\section{Introduction}\label{sec:intro}

Many compelling applications of machine learning involve the collection and processing of sensitive personal data, giving rise to concerns about privacy~\cite{AtenieseFMSVV13,cormode-bayes,mi2015,fredrikson2014privacy,mitheory2016,Li2013,shokri16shadow,nasr18whitebox,dptheory2016,brickell-utility}. 
In particular, when machine learning algorithms are applied to private training data, the resulting models might unwittingly leak information about that data through their behavior or representation.


Membership inference (MI) attacks aim to determine whether a given data point was present in the training set used to build a model. 
This can be a privacy threat in itself, but vulnerability to MI has also come to be seen as a more general indicator of whether a model leaks private information~\cite{long17,shokri16shadow,yeom17overfitting}, and is closely related to the guarantee provided by differential privacy~\cite{Li2013}.

To date, most MI attacks follow the so-called \emph{shadow model} approach~\cite{shokri16shadow}.
This approach casts the attack as a supervised learning problem, where the adversary is given a data point and its true label, and aims to predict a binary label indicating membership status.
To do so, the adversary trains a set of \emph{shadow models} to replicate the functionality of the target model, and trains an \emph{attack model} from data derived from the shadow models' outputs on the points used to train each shadow model and points not previously seen by each shadow model.

Subsequently, \citeauthor{nasr18whitebox} extended this attack to the white-box setting~\cite{nasr18whitebox} by including activation and gradient information obtained from the target model as features for the attack model.
However, \citeauthor{nasr18whitebox} find that a simple extension of the shadow model approach to the white-box setting does not produce an effective attack~\cite{nasr18whitebox} (we discuss why in Section~\ref{sec:deep}); thus, their white-box attack deviates from the threat model common to most work on MI, and instead assumes that the adversary \emph{already knows a significant portion of the target model's training data}.
Features to train the attack model are obtained directly from the target model, using the gradients, activations, and outputs obtained by evaluating on known member/non-member points.
\emph{In this paper, we present an effective white-box MI attack that operates without access to \emph{any} of the target model's training data.}
Crucially, our analysis uncovers a more intimate understanding of how overfitting takes place in a model, which we leverage to create our attack.

\paragraph{Finding Evidence of Membership}

In this paper, we take a fresh look at the problem of white-box membership inference.
We begin with the intuitive observation that while overfitting leads to privacy issues because the model ``memorizes'' certain aspects of the training data, this is not necessarily manifested in the model's output behavior.
Instead, \emph{it is likely to show up in the way that the model uses features}---both those that are given explicitly and that are learned in internal layers.

\begin{figure}
	\centering
	\subfloat[\ ]{\label{fig:train_samples}
		\resizebox{\columnwidth}{!}{%
			\includegraphics{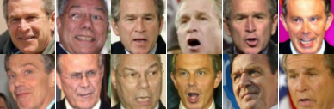}
		}
	}

	\subfloat[\ ]{\label{fig:test_expls}
		\resizebox{.72\columnwidth}{!}{%
			\includegraphics{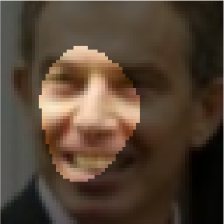}
			\includegraphics{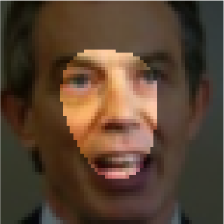}
			\includegraphics{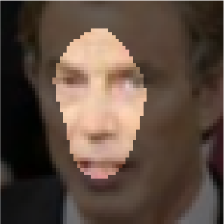}
		}
	}
	\subfloat[\ ]{\label{fig:train_expls}
		\resizebox{.24\columnwidth}{!}{%
			\includegraphics{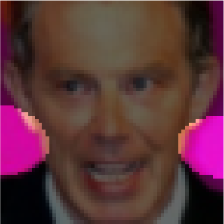}
		}
	}

	\caption{\label{fig:pictorial_example}
		Pictorial example of how overfitting can lead to idiosyncratic use of features. (a) shows 12 training instances. We see that the image of Tony Blair on the top right has a distinctive pink background. (b) depicts internal explanations~\cite{leino18influence} for three test instances. The explanations show that the model uses Tony Blair's face to classify these instances, as we might expect. Meanwhile, (c) shows the explanation for the image with the distinctive pink background from the training set, where we see that the model is using the pink background to infer that the image is of Tony Blair.
	}
\end{figure}

Intuitively, we posit that idiosyncratic features present in the training data, which are predictive \emph{only} for the training data but not the sampling distribution, are oftentimes encoded in the model during training.
Consider the example illustrated by Figure~\ref{fig:pictorial_example}, in which a model was trained to recognize faces from the \emph{Labeled Faces in the Wild} (LFW) dataset.
Figure~\ref{fig:train_samples} shows 12 instances sampled from the training set of the model.
The top right corner of Figure~\ref{fig:train_samples} depicts an image of Tony Blair with a distinctive pink background.
Supposing that the background is unique to this training instance, an overfit model may use the background as a feature for classifying Tony Blair, identifying the instance as a member of the training set via the uncharacteristic way in which the model correctly labels it.
In such a setting, the model's use of the pink background could be viewed as evidence of membership.

Figures \ref{fig:test_expls} and \ref{fig:train_expls} show this phenomenon on a convolutional neural network trained on this dataset. Figures \ref{fig:test_expls} and \ref{fig:train_expls} visualize the regions of the image most influential~\cite{leino18influence} towards the classification of ``Tony Blair'' on three test instances, and on the aforementioned training instance with the pink background.
While the model is influenced most by Tony Blair's face for classification on the test instances, on the training instance it relies on the distinctive pink background.


We show that this evidence-based approach can be used on a variety of real datasets to infer membership, and leverage it to develop a new attack (Sections~\ref{sec:analysis} and \ref{sec:deep}) that outperforms previous attacks (Section~\ref{sec:eval}).



\paragraph{Calibrating Confidence}
By far the simplest MI attack, which we dub the ``naive'' attack, follows from the fact that generalization error necessarily leads to membership vulnerability~\cite{yeom17overfitting}.
Given a data point and its true label, the attacker runs the model and observes whether its predicted label is correct.
If it is, then the attacker concludes that the point was in the training data; otherwise, the point is presumed a non-member.
Surprisingly, \emph{in many cases this works as well as the shadow model attack} (Section~\ref{sec:eval_comparison}, Figure~\ref{fig:all_apr}).
As a practical attack, the naive method has a significant drawback even when it appears yield reasonable accuracy.
Namely, it does not provide the attacker with much \emph{confidence} about a positive inference: the point may have been a training set member, or it may just have been classified correctly.
After all, this is how the model is intended to behave on test points, so it may not be sensible to base a membership inference on a correct prediction result.

Initially, it may seem that shadow model attacks do not inherit this limitation, as the attack model can be trained to emit a confidence score with its prediction.
If this score is well-calibrated, then an attacker could use it to make more confident inferences.
Unfortunately, we find shadow attacks are not typically well-calibrated; in fact, Figure~\ref{fig:rule_of_thumb} (Section~\ref{sec:eval_precision}) shows that raising the confidence threshold for positive prediction sometimes \emph{decreases} the precision of the attack.
In short, like the naive attack, the shadow model attack often produces little consistently useful information to characterize the likelihood that a positive inference is correct.

\emph{We posit that if the adversary \textbf{confidently} identifies even one training point, then it is reasonable to say that a privacy violation occurred.}
We therefore propose that an effective attack should have the ability to make confident inferences, underscoring the need for attacks with high precision.
To this end, we demonstrate that the confidence scores accompanying the inferences made by our attack can be used to accurately calibrate its precision (Section~\ref{sec:eval_precision}, Figure~\ref{fig:rule_of_thumb}).

\paragraph{Evaluating Defenses}
A number of defenses have been proposed for membership inference.
\emph{Differential privacy} (DP)~\cite{dwork06}, in addition to regularization methods like dropout~\cite{srivastava14} in deep nets are two commonly-proposed defenses.
While differential privacy gives a theoretical guarantee against membership inference~\cite{yeom17overfitting}, a \emph{meaningful} guarantee---one that bounds the probability of attack success below 1---requires an $\epsilon$ that is considerably smaller than what is often used in practice.
Nonetheless, common wisdom conjectures that large-$\epsilon$-DP may provide a practical defense, particularly if the privacy budget analysis only gives a loose bound on $\epsilon$.

Unfortunately, we find that this is not necessarily the case.
We test our attack on deep models trained with $(\epsilon, \delta)$-differential privacy using the moments accountant method~\cite{abadi16} (Section~\ref{sec:defenses}), and find that training with a large $\epsilon$ sometimes provides little defense against our attack when compared against its effectiveness on non-private models.
These results demonstrate that practical MI attacks like the one described in this paper can serve as a heuristic measure to evaluate paramater choices in private learning, while also emphasizing the need for more research in this area.

\paragraph{Organization}
In Section~\ref{sec:background}, we introduce background on membership inference and machine learning. 
Section~\ref{sec:analysis} describes the evidence-based attack, beginning in an idealized setting that can be rigorously analyzed  to motivate the intuition behind the attack (Section~\ref{sec:bayes-opt-mi}).
Subsequently, we gradually lift the generative assumptions used in this derivation to obtain an attack that works well on real data (Sections \ref{sec:categorical} and \ref{sec:generalizing}).
Section~\ref{sec:calibrating} discusses calibration, and
Section~\ref{sec:deep} shows how our attack can be extended to deep networks.
Section~\ref{sec:eval} presents our evaluation on both synthetic data and nine real datasets derived from real-world medical and financial data, and common benchmark datasets.
Section~\ref{sec:defenses} discusses defenses against MI attacks and tests their efficacy against our attack.
Section~\ref{sec:related} covers related work, and
Section~\ref{sec:future} concludes the paper.

\section{Background}\label{sec:background}

Membership inference (MI) attacks aim to determine whether a given data point was present in the dataset used to train a given target model.
In this section, we begin by introducing the necessary background needed to formally define membership inference, as well as explicitly defining the threat model used in our analysis.

\subsection{Supervised Learning and Target Models}
We assume data from some universe $\mathcal U = \mathcal X~\times~\mathcal Y \subset \mathbb R^n \times [C]$, drawn from a distribution, $\distrib^*$.
Consistent with the typical supervised learning setting, $x\in\mathcal X$ is a vector of $n$ features and $y\in\mathcal Y$ is a label or classification target, corresponding to $C$ distinct classes.
Given a loss function, $\mathcal{L} : \mathcal{X}\times\mathcal{Y} \to \mathbb{R}$, the goal of supervised learning is to construct a model, $g$, that minimizes $\mathcal{L}(g(x), y)$ on future unseen samples, $x$, drawn from $\distrib^*$.
This is achieved by minimizing $\mathcal{L}(g(x), y)$ on a finite training set, $S$, drawn i.i.d. from $\distrib^*$.

A membership inference attack operates on a particular \emph{target model}, $\hat g$.
In this work, we consider target models that are expressed as feed-forward neural networks; i.e., they consist of successive linear transformations, or layers, where each layer, $\ell$, is parameterized by a matrix of weights and biases $\mathcal{W}_\ell$, $\mathcal{B}_\ell$, followed by the application of a non-linear activation function. 

Consistent with common practice, we assume that internal layers use the rectified-linear (ReLU) activation: $\mathit{relu}(x) = \mathrm{max}(0, x)$.
We assume that the final layer has one component for each label in $[C]$ and uses the softmax activation: $\softmax(x)_j = e^{x_j}/\sum_i{e^{x_i}}$.
The use of the softmax function is standard in machine learning for multi-class classification.
Models trained in this way produce \emph{confidence scores} for each label that can be interpreted as probabilities~\cite{Goodfellow2016}.

In the simplest case we consider, the target model consists of a single layer with only the softmax activation, and is a \emph{linear softmax regression} model.
We will sometimes refer to this type of model by its parameterization, $\hat W$, $\hat b$.
Our approach generalizes to \emph{deep networks} where the target model has multiple successive internal ReLU-activated layers, followed by a single softmax output layer.


\subsection{Membership Inference}
We adpot a formulation of Membership Inference attacks similar to that of \citeauthor{yeom17overfitting}~\cite{yeom17overfitting}.
First a value, $b$, is chosen uniformly at random from $\{0, 1\}$.
If $b = 1$, the attacker, $\mathcal{A}$, is then given an instance $(x,y)$ from the general population; otherwise, if $b = 0$, $(x,y)$ is sampled uniformly at random from the elements of the training set, $S$, used to generate target model, $\hat g$.
The attacker then attempts to predict $b$ given $(x,y)$ and some additional knowledge, $\mathit{aux}(\hat g)$, about $\hat g$ determined by the threat model (see below).

\paragraph{Threat Model}
\label{sect:threat-model}
Prior work~\cite{shokri16shadow,yeom17overfitting} has focused primarily on the so-called \emph{black-box} model where the adversary has access to $\distrib^*$, the learning algorithm used to produce $\hat g$ (including hyperparameters), the size of the training set, and the ability to query $\hat g$ arbitrarily on new points.
In practice, having access to $\distrib^*$ amounts to knowing a finite data set, $\tilde S$ (distinct from $S$), sampled i.i.d. from $\distrib^*$.

In this work, we replace black-box access to $\hat g$ with \emph{white-box} access.
Rather than only being able to query the target model, the attacker has access to the exact representation of $\hat g$ that was produced by the learning algorithm and used by the model owner to make inferences on new data.
For the target models commonly used in practice, e.g. neural networks and linear classifiers, this amounts to a set of floating-point weight matrices and biases, in addition to the linear operators and activation functions used at each layer.

This threat model reflects the growing number of publicly-available models on websites like Model Zoo~\cite{modelzoo}, as well as the fact that white box representations may fall into the hands of an adversary via other means (e.g., a security breach). 
Additionally, even in situations where the requirements for a white-box attack may not be practical for an adversary, the ability to mount a more powerful attack could be useful for a defender, as it provides a more conservative estimate of the potential threat.

\paragraph{Metrics}
The \emph{accuracy} of an attack is the probability that $\mathcal{A}$'s prediction is equal to $b$, taken over the randomness of $b$, $(x,y)$, and $\mathcal{A}$.
Because an adversary that guesses randomly achieves $50\%$ accuracy, we will often opt to describe the \emph{advantage} of an attack~\cite{yeom17overfitting}, given by Equation~\ref{eq:advantage} in terms of attack, $\mathcal{A}$. Advantage scales accuracy to the $50\%$ baseline to yield a measure between -1 and 1.
\begin{equation}
\label{eq:advantage}
\mathrm{advantage}(\mathcal{A}) = 2\mathrm{Pr}\left[\mathcal{A}\big((x,y),~\mathit{aux}(\hat g)\big) = b\right] - 1
\end{equation}
While advantage is an indicator of the degree to which private information is leaked by the model, it does not necessarily capture the severity of the threat posed to any given individual in the training set.
From this perspective, a privacy violation occurs if \emph{any} of the points can be confidently identified by the adversary---this is arguably a greater threat than if the adversary were to identify every training member with very low confidence.
Thus, we also consider \emph{precision} (Equation~\ref{eq:precision}) as a key desideratum for the attacker. 
In order for an attacker to reach confident inferences, precision must be appreciably greater than 1/2.
If no points are predicted to be members, we define precision to be 1/2.
\begin{equation}
\label{eq:precision}
\mathrm{precision}(\mathcal{A}) = \mathrm{Pr}\left[b = 1 | \mathcal{A}\big((x,y),~\mathit{aux}(\hat g)\big) = 1\right]
\end{equation}
Finally, we include \emph{recall} (Equation~\ref{eq:recall}) as a metric in our evaluation as it has been reported in prior work.
However, we place less emphasis on this metric, as an attack with high recall is not necessarily effective in practice if it fails to return confident inferences on any points.
For example, an adversary that simply predicts that \emph{all} points are members achieves perfect recall, yet this clearly does not constitute a practical attack.
\begin{equation}
\label{eq:recall}
\mathrm{recall}(\mathcal{A}) = \mathrm{Pr}\left[\mathcal{A}\big((x,y),~\mathit{aux}(\hat g)\big) = 1 | b = 1\right]
\end{equation}

\paragraph{Logistic Attack Models}
In the interest of achieving good precision, we consider attacks that yield confidence scores with their predictions.
Thus, we can think of membership inference as a binary logistic regression~\cite{Murphy2012} problem, in which a logistic (\emph{sigmoid}) function models confidence with respect to the binary dependent variable (i.e., membership or non-membership).
The sigmoid function, $\sigmoidf$, is is given by $\sigmoid{x} = \frac{1}{1 + e^{-x}}$, and can be thought of as converting the log-odds of the dependent variable to a probability.
The use of the sigmoid function for binary classification is standard in machine learning, and has been applied in prior membership inference attacks as well~\cite{shokri16shadow}.

\section{White-box Membership Inference}\label{sec:analysis}

In this section, we introduce our core membership inference attack.
Starting in an idealized setting where the exact data distribution is known and the model is linear, we proceed by deriving the Bayes-optimal logistic attack model (Section~\ref{sec:bayes-opt-mi}).
We show that when the data-generating assumptions hold, the confidence scores produced by this attack correspond to the true membership probability, and can thus be used for effective, accurate calibration towards high-precision attacks.
Using the insights gained from this analysis, we then show how to generalize the attack to settings where the data-generating distribution is unknown or does not match our theoretical assumptions (Sections~\ref{sec:categorical} and \ref{sec:generalizing}), and discuss calibration in this setting (Section~\ref{sec:calibrating}).
In Section~\ref{sec:deep} we extend the attack to deep models.

\subsection{Overview of the attack}\label{sec:bayes-opt-mi}

Our attack works from the intuition that when models overfit to their training data, they potentially leak membership information through anomalous behavior at test time.
However, while this behavior may manifest itself in the form of prediction errors on unseen points, this need not be the case, and a more nuanced look at how memorization occurs yields new insights that can be used in an attack.

Models use features to distinguish between classes, and while some features may be truly discriminative (i.e., function as good predictors on unseen data), others may be discriminative only on the particular training set merely by coincidence.
When the model applies features of the latter type to make a prediction, this can be thought of as ``evidence'' of overfitting regardless of whether the prediction is correct; the salience of a feature coincidental to the training data is suggestive on its own.
Similarly, there may be features that are discriminative on the data in general, but not on the training data.

For example, consider a hypothetical model trained to recognize celebrity faces.
Suppose that in reality, each celebrity is wearing sunglasses in $10\%$ of his or her respective pictures, so the presence of sunglasses is not an informative feature for this task.
However, if the training data used to construct the model contained images of a particular subject wearing sunglasses with greater frequency, say $30\%$, then the model might learn a feature that detects sunglasses in an internal layer, and weight this feature towards prediction of that subject.
Knowing that the presence of sunglasses is not predictive of identity on the true distribution, an attacker would infer that, all else being equal, a picture of this subject wearing sunglasses is more likely to be a training set member.

While this may not be conclusive evidence of membership, it can be aggregated with other aspects of the model's behavior on an instance to make a final determination with greater confidence than would be possible using only black-box information. 
To see why this is the case, consider that another model trained on a different sample, e.g. one that reflects a ``normal'' frequency of subjects wearing sunglasses, may learn to make the same numerical predictions using a different set of features. 
A black-box attacker would be unable to distinguish these cases, and thus be deprived of the feature-based evidence available through an examination of the model's use of internal features.

This example highlights the intuition that membership information is leaked via a target model's idiosyncratic use of features. Essentially, features that are distributed differently in the training data from how they are distributed in the true distribution can provide evidence either for or against membership.
Our attack works by deriving a set of parameters that profile idiosyncratic feature use, which are then used to construct a logistic attack model.


\subsection{A Bayes-Optimal Attack}\label{sec:bayes-opt-mi}

To motivate this intuition more formally, we begin by showing how to mount this evidence-based attack in an idealized setting where data is distributed according to a known distrubution. 
This provides a simpler illustration of the central ideas used in our later attack, \emph{where we do not make explicit assumptions about the data distribution}.
We show that the attack in this setting leads to Bayes-optimal membership predictions on points from that distribution, which suggests that even when the strict assumptions made here are violated, the approach may nonetheless be a strong heuristic even if it cannot be proved optimal.

\paragraph{Generative Assumptions}
Recall the setting described in Section~\ref{sec:background}: a model, $\hat g$, trained on $S\sim\distrib^*$, and an adversary that leverages white-box access to $\hat g$ to create an attack model, $m$, that predicts whether an instance, $(x,y)\in\mathcal U$, belongs to $S$. 
We show how the example above can be extended to this setting by introducing some assumptions about $\hat g$ and $\distrib^*$.

First we assume that $\distrib^*$ is given by parameters, $\mu^*_y$, $\Sigma^*$, and $p^* = (p^*_1, \ldots, p^*_C)$, such that the labels, $y$, are distributed according to a Categorical distribution with parameter $p^*$, and the features, $x$, are multivariate Gaussians with mean $\mu^*_y$ for each label $y$, and covariance matrix, $\Sigma^*$. 
\begin{equation}
\label{eq:nb-distr}
y \sim \mathrm{Categorical}(p^*)\quad x \sim \N(\mu^*_y, \Sigma^*)
\end{equation}
Furthermore, assume that $\Sigma^*$ is a diagonal matrix, i.e., the distribution of $x$ satisfies the naive-Bayes assumption of the features being independent conditioned on the class.
We will therefore write $\Sigma^*_{jj}$ as $\sigma^{*2}_j$.

Recall that $S$ is drawn i.i.d. from $\distrib^*$, so its samples are also distributed according to Equation~\ref{eq:nb-distr}.
However, the empirical means and variance of $S$ will not match those of $\distrib^*$ exactly, except in expectation.
Therefore, we denote by $\hat\distrib$ the \emph{empirical distribution} of the training data, $S$. 
Let $\hat p$ be the empirical class prior for $S$, $\hat\mu_y$ be the empirical mean of the features in $S$ with class $y$, and $\hat\Sigma$ be the empirical covariance matrix of the features in $S$.
We make the analogous assumption that $\hat\Sigma$ is a diagonal matrix, and that the empirical distribution function can be modeled as a normal distribution, $\N(\hat\mu, \hat\Sigma)$.
Intuitively, we can now think of $m$ as determining whether $(x, y)$ is more likely to have been drawn from $\hat\distrib$ (i.e., $(x,y)\in S$), or $\distrib^*$.

If we momentarily assume that the attacker knows $\distrib^*$ and $\hat\distrib$, then we can proceed to derive an attack model purely in terms of their respective parameters, namely $\mu_y^*$, $\hat\mu_y$, $\Sigma^*$, and $\hat\Sigma$.

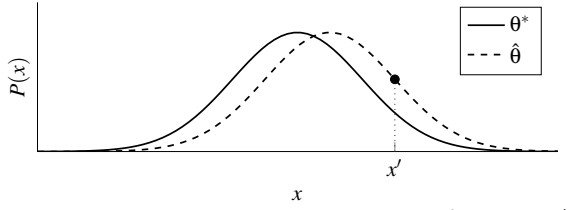
\begin{figure}
    \centering
    \resizebox{0.85\columnwidth}{!}{%
    \hspace*{-1.5em}
    \pgfplotsset{compat=1.13}
\makeatletter
\pgfmathdeclarefunction{normaldist}{3}{%
	\begingroup
	\pgfmathparse{1 / sqrt(2*3.14159*#3^2) * exp(-(#1 - #2)^2 / (2*#3^2))}%
	\pgfmath@smuggleone{\pgfmathresult}%
	\endgroup
}
\makeatother

\centering
\begin{tikzpicture}
	\begin{axis}[
		axis x line*=bottom,
		axis y line*=left,
		no markers,
		ylabel={$P(x)$},
		xlabel={$x$},
		xmin=-4, xmax=4,
		ymin=0, ymax=0.5,
		ytick=\empty,
		xtick=\empty,
		extra x ticks={1.5},
		extra x tick labels={$x'$},
		domain=-5:5,
		samples=100,
		legend pos=north east,
		legend cell align=left,
		width=10cm,
		height=4cm		
	]
	\addplot[thick]{normaldist(x,0.0, 1)};
	\addplot[thick, dashed]{normaldist(x,0.5, 1)};
	\draw [dotted] (1.5,0) -- (1.5, 0.24197082671);
	\addplot[only marks,samples at={1.5}]{normaldist(x,0.5,1)};

	\legend{$\theta^*$, $\hat\theta$}
	\end{axis}
\end{tikzpicture}}
    \vspace*{-1em}
    \caption{\label{fig:theta_star_hat}
        Example of two Gaussian distributions, $\eta^*$ and $\hat\eta$. The point $x'$ has a higher probability of being generated by $\hat\eta$ than by $\eta^*$. Given a prior probability of $\half$ for being drawn from either distribution, the decision boundary for predicting which distribution a given point was drawn from would be at the intersection of the two curves, and $x'$ would be predicted to have been drawn from $\hat\eta$.
    }
\end{figure}

\paragraph{Attack Model}
Consider two Gaussian distributions, $\eta^* = \N(\mu^*, \sigma^*)$ and $\hat\eta = \N(\hat\mu, \hat\sigma)$.
For $x \in \mathbb R$, $x$ is more likely to have been generated by $\hat\eta$ than by $\eta^*$ when $\N(x\ |\ \hat\mu, \hat\sigma) > \N(x\ |\ \mu^*, \sigma^*)$.
An example of this is shown pictorially in Figure~\ref{fig:theta_star_hat}.
Assuming a prior probability of $1/2$ for being drawn from either distribution, we could construct a simple model that predicts whether $x$ was drawn from $\hat\eta$ rather than $\eta^*$ by solving for $x$ in this inequality. When the variances, $\sigma^*$ and $\hat\sigma$, are the same, this produces a linear decision boundary as a function of $\mu^* - \hat\mu$ and $\sigma^*$.

Our setting is more complicated than this simple Gaussian example, but as we demonstrate below, the same principle can be applied to mount an attack.
Let $(X, Y)$ be random variables drawn from either $\hat\distrib$ or $\distrib^*$ (as defined above), with probability $t$ of drawing from $\hat\distrib$.
Let $T$ be the event $(X, Y)\in S$, i.e., that a point drawn according to this process was in the training set. 
Thus, $\p{T} = t$.
In keeping with the MI definition presented in Section~\ref{sec:background}, we will assume that $t = \half$.
We want an attack model, $m^y(x)$, to give us the probability that point $(x, y)$ is a member of the training set, $S$.

Because we know $t$ and the parameters of $\distrib^*$ and $\hat\distrib$, we can derive an estimator for this quantity by applying Bayes' rule and algebraically manipulating the result to fit a logistic function of the log odds.
We then make use of the naive-Bayes assumption, allowing us to write the probability of observing $x$ given its label as the product of the probabilities of observing each of $x$'s features independently. The result is linear in the target feature values when $\hat\sigma = \sigma^*$, as detailed in Theorem~\ref{thm:bayes-opt}.
The proof for Theorem~\ref{thm:bayes-opt} is given in Appendix~\ref{app:proof}.

\begin{theorem}
\label{thm:bayes-opt}
Let $x$ and $y$ be distributed according to $\distrib^*$, given by Equation~\ref{eq:nb-distr} with parameters $(p^{*}, \mu_y^*, \Sigma^*)$, and $S$ be drawn i.i.d. from $\distrib^*$, with empirical distribution function, $\hat\distrib$, modeled as $y'\in S\sim\mathrm{Categorical}(\hat p)$, $x'\in S\sim\N(\hat\mu_{y'}, \hat\Sigma)$.
Further, assume that $\hat\Sigma = \Sigma^*$ is diagonal and $\hat p = p^*$.
Then the Bayes-optimal predictor for membership is given by Equation~\ref{eq:bayes-predictor}.
\begin{align}
\label{eq:bayes-predictor}
& m^y(x) = \sigmoid{{w^y}^Tx + b^y} \\[4pt]
\nonumber\mathrm{where\qquad } & w^y = \frac{\hat\mu_{y} - \mu^*_{y}}{\sigma^2}\quad
b^y = \sum_j{\frac{\mu_{yj}^{*2} - \hat\mu_{yj}^2}{2\sigma_j^2}}
\end{align}

\end{theorem}
Notice that the magnitude of the attack model weights given in Theorem~\ref{thm:bayes-opt} is large only on features whose mean on the training data differs significantly from its mean in the distribution, $\distrib^*$, relative to that feature's variance.
This is a manifestation of the intuition described in the previous section, as the attack model effectively treats those features as its primary  ``evidence'' for deciding membership.
We also point out that the attack model detailed in Theorem~\ref{thm:bayes-opt} defines a different set of parameters for each class label, $y$.
This follows from the generative assumptions, as each class may have a distinct mean, and thus must be distinguished using separate critera.
As a practical matter this is not an impediment, as our setting assumes that the true class label is given to the adversary, so there is no ambiguity as to which set of parameters should be applied.

\paragraph{Summary}
Features that are more likely in the empirical training distribution, $\hat\distrib$, than in the true ``general population'' distribution, $\distrib^*$, serve as evidence for membership.
Theorem~\ref{thm:bayes-opt} shows how this evidence can be compiled into a linear attack model, $w^y, b^y$, that achieves Bayes-optimality for membership inference when both distributions are known precisely.
In Section~\ref{sec:categorical}, we show how to obtain approximate values for $w^y$ and $b^y$ when the distributions are unknown.

\subsection{Obtaining MI Parameters from Proxy Models}\label{sec:categorical}

In practice, it is unrealistic to know the exact parameters defining the distributions $\distrib^*$ and $\hat\distrib$. 
In particular, our threat model assumes that the attacker has no \textit{a priori} knowledge of the parameters of $\hat\distrib$ or the elements of $S$, only that $S$ was drawn from $\distrib^*$.
While we assume white-box access to the target model, $\hat g$, we cannot expect that it will explicitly model $\hat\distrib$; indeed, $\hat g$ is usually parameterized by weights, leaving the distribution parameters underdetermined.
Finally, $\distrib^*$ and $\hat\distrib$ may violate the naive-Bayes assumption, or be difficult to parameterize directly.

These issues can be largely addressed by observing that the learned weights are sensitive to $\hat\distrib$, and although they may not encode sufficient information to solve for the exact parameters, they may encode useful information about the differences between $\hat\distrib$ and $\distrib^*$.
To measure these differences, we use a \emph{proxy dataset}, $\tilde S$, which is drawn i.i.d. from $\distrib^*$ (but distinct from $S$) to train a proxy model, $\tilde g$, which is then compared with $\hat g$.
To control for differences in the learned weights resulting from the learning algorithm, rather than from differences between $\hat\distrib$ and $\distrib^*$, the proxy model is trained using the same algorithm and hyperparameters as $\hat g$ (note that this information is assumed to be known in our threat model).
This process can be repeated on many different $\tilde S$, using bootstrap sampling when the available data is limited.

In more detail, we continue with the assumption that data is generated according to Equation~\ref{eq:nb-distr}. 
Note that our target is a linear model, $\hat W, \hat b$, that minimizes 0-1 loss on $S$ for the predictions given by $\argmax_{c\in[C]}\{\softmax(\hat W^Tx + \hat b)_c\}$.
This is a convex optimization problem that, under our generative assumptions, is minimized when $\hat W$ and $\hat b$ are given by Equation~\ref{eq:bayes_opt_weights}\footnote{see \citeauthor{murphy-lecture}, Slide 20~\cite{murphy-lecture} for details.}.
\begin{equation}\label{eq:bayes_opt_weights}
    \hat W_{jy} = \frac{\hat\mu_{yj}}{\hat\sigma_j^2} \qquad 
    \hat b_{y} = \sum_j{\frac{-\hat\mu_{yj}^2}{2\hat\sigma_j^2}} + \log(\hat p)
\end{equation}
Plugging this, and the analogous equation for the proxy model, $\tilde W, \tilde b$, into Equation~\ref{eq:bayes-predictor} from Theorem~\ref{thm:bayes-opt}, we see that the weights and biases of the attack model $m^y$ are approximated by $w^y \approx \hat W_{:y} - \tilde W_{:y}$ and $b^y \approx \hat b_y - \tilde b_y$ respectively, assuming that $\tilde\mu \approx \mu^*$.
This is summarized in Observation~\ref{thm:opt_wb}, which leads to a natural attack as shown in Algorithm~\ref{alg:bayes}. We call this the \bayesian attack.
\begin{obs}\label{thm:opt_wb}
    For linear softmax model, $\hat g$, with weights, $\hat W$, and biases, $\hat b$; and proxy model, $\tilde g$, with with weights, $\tilde W$, and biases, $\tilde b$, the Bayes-optimal membership inference model, $m$, on data satisfying Eq.~\ref{eq:nb-distr} is approximately
    \begin{align}
        \label{eq:opt_w_b}& m^y = \sigmoid{{w^y}^T x + b^y} \\[4pt]
        \nonumber\mathrm{where\qquad} & 
        w^y = \hat W_{:y} - \tilde W_{:y}\quad
        b^y = \hat b_y - \tilde b_y 
    \end{align}
\end{obs}
\begin{algorithm}[t] 
\small
\Fn{\FmakeModel($\hat g$, $\tilde S$)}{
    $\tilde g\ \leftarrow\ \FtrainProxy(\tilde S)$\;
    $w^y\ \leftarrow\ \hat g.W_{:y} - \tilde g.W_{:y}\qquad \forall y\in[C]$\;
    $b^y\ \leftarrow\ \hat g.b_{:y} - \tilde g.b_{:y}\qquad \forall y\in[C]$\;
    \KwRet $\lambda (x,y) : \sigmoid{{w^y}^T x + b^y}$\;
}
\Fn{\Fpredict($m$, $x$, $y$)}{
    \KwRet $1$ if $m^y(x) > \half$ else $0$\;
}
\caption{\label{alg:bayes} The Linear \bayesian MI Attack}
\end{algorithm}
Notice that Observation~\ref{thm:opt_wb} gives the weights and biases of $m^y$ in terms of only the observable parameters of the target and proxy models.
This is therefore possible \emph{even when the distributions, $\distrib^*$ and $\hat\distrib$, are unknown}.
Furthermore, while Observation~\ref{thm:opt_wb} is derived and stated using relatively strong generative assumptions, we find in Section~\ref{sec:eval} that this attack is nevertheless often effective when these assumptions do not hold. 
In Section~\ref{sec:generalizing} we show how to further relax these generative assumptions.

\subsection{Learning to Generalize to Arbitrary Distributions}\label{sec:generalizing}

One way of viewing the \bayesian attack is that it weights membership predictions by measuring a sort of displacement between the weights of the target model and the ideal weights of the true distribution as approximated by the proxy model.
Let $d_f : \mathbb{R}\times\mathbb{R} \rightarrow \mathbb{R}$ be a \emph{displacement function} that is applied element-wise to the weights of the model --- for vectors $x$ and $y$, let $D(x, y) = (d_f(x_1, y_1), \ldots, d_f(x_n, y_n))$.
We can express the \bayesian attack via a such a displacement function, namely, $w^y = D(\hat W_{:y}, \tilde W_{:y})$ and $b^y = D(\hat b_y, \tilde b_y)$, by letting $d_f(x, y) = x - y$, i.e., by setting $D$ to be element-wise subtraction.

As per Observation~\ref{thm:opt_wb}, element-wise subtraction is optimal for membership inference under the Gaussian naive-Bayes assumption, but it may be that for other distributions, a different displacement function is more appropriate.
More generally, we can represent the displacement function as a neural network, and train it using whatever data is at hand.

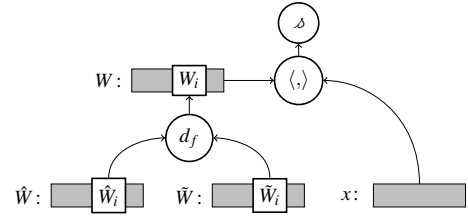
\begin{figure}
    \centering
    \resizebox{!}{8em}{

\begin{tikzpicture}[
		roundnode/.style={
			circle, 
			draw=black,
			thick, 
			minimum size=7mm},
		squarednode/.style={
			rectangle, 
			draw=black, 
			thick, 
			minimum size=5mm},
		]



	\node (whl) at (0,.2) {$\hat W:$};
	\draw[fill=lightgray] (.4,0) rectangle (2.0,.4);
	\node[squarednode, fill=white] (whi) at (1.4,.2) {$\hat W_i$};
	\node (wh) at (.8,.4) {};

	\node (wtl) at (2.8,.2) {$\tilde W:$};
	\draw[fill=lightgray] (3.2,0) rectangle (4.8,.4);
	\node[squarednode, fill=white] (wti) at (4.2,.2) {$\tilde W_i$};
	\node (wt) at (3.6,.4) {};

	\node (xl) at (5.6,.2) {$x:$};
	\draw[fill=lightgray] (6.0,0) rectangle (7.6,.4);




	\node[roundnode] (f) at (2.8, 1.2) {$d_f$};
	\draw[->] (whi) .. controls +(up:8mm) and +(left:8mm) .. (f);
	\draw[->] (wti) .. controls +(up:8mm) and +(right:8mm) .. (f);

	\node (wl) at (1.4, 2.2) {$W:$};
	\draw[fill=lightgray] (1.8, 2.0) rectangle (3.4, 2.4);
	\node[squarednode, fill=white] (wi) at (2.8, 2.2) {$W_i$};
	\draw[->] (f) -- (wi);

	\node[roundnode] (dp) at (4.7, 2.2) {$\langle,\rangle$};
	\draw[->] (3.4, 2.2) -- (dp);
	\draw[->] (6.8, .4) .. controls +(up:12mm) and +(right:12mm) .. (dp);

	\node[roundnode] (s) at (4.7, 3.2) {$\sigmoidf$};
	\draw[->] (dp) -- (s);


\end{tikzpicture}}
    \caption{\label{fig:generalized_approach}
        Illustration of the generalized attack model.
        A learned displacement function, $d$, is applied element-wise to the weights of the target and proxy model to produce attack model weights, $W$.
        The inner product of $W$ and $x$ is then used to make the membership prediction. 
        \emph{Not pictured: d is also applied to the biases, $\hat b$ and $\tilde b$, to produce $b$, which is added to the result of the inner product.}
    }
\end{figure}

Figure~\ref{fig:generalized_approach} illustrates this approach, which we call the \general attack.
A learned displacement function, $d_f$, is applied element-wise to $\hat W$ and $\tilde W$ to produce attack model weights, $W$, and to $\hat b$ and $\tilde b$ to produce attack model biases, $b$.
It then predicts the probability of membership as $\sigmoid{W_{:y}^T x + b_y}$.

As $d_f$ is applied element-wise to pairs of weights, we model $D$ as a 1-dimensional convolutional neural network, where the initial layer has a kernel size and strides of 2 (i.e., the kernel is applied to one element of $\hat W_{:y}$ and one element of $\tilde W_{:y}$), and subsequent layers have a kernel size and stride of 1.

\begin{algorithm}[t] 
\small
\Fn{\FmakeModel($\hat g$, $\tilde S$, $N$)}{
    \For{$i\in[N]$}{
        $\tilde S^{1}_i, \tilde S^{0}_i\ \leftarrow\ \Fsplit_i(\tilde S)$\;
        $\check g_i\ \leftarrow\ \FtrainShadow(\tilde S^{1}_i)$\;
        $\tilde g_i\ \leftarrow\ \FtrainProxy(\tilde S^{0}_i)$
    }
    $T\leftarrow 
        \underset{\forall (x, y')\in \tilde S^\ell_i : y' = y,\ \ 
            \forall y\in[C],\ \ 
            \forall \ell \in \{0, 1\},\ \ 
            \forall i\in[N]}
        {\left[(\check g_i.W_{:y},\ \tilde g_i.W_{:y},\ \check g_i.b_y,\ \tilde g_i.b_y,\ x,\ \ell)\right]}
    $\;
    $D \leftarrow{\scriptstyle\underset{D'}{\text{argmin}}}\left\{
    \underset{\scriptscriptstyle(\hat w, \tilde w, \hat b, \tilde b, x, \ell)\in T}{\mathbb{E}}
    \left[
        \mathcal{L}(
            {\scriptstyle\sigmoid{D'(\hat w, \tilde w)^T x + D'(\hat b, \tilde b)}, \ell})
        \right]
    \right\}$\;
    $\tilde g \leftarrow \FtrainProxy(\tilde S)$\;
    \KwRet {\small$\lambda (x,y) : \sigmoid{D(\hat g.W_{:y}, \tilde g.W_{:y})^T x + D(\hat g.b_y, \tilde g.b_y)}$}\;
}
\Fn{\Fpredict($m$, $x$, $y$)}{
    \KwRet $1$ if $m^y(x) > \half$ else $0$\;
}
\caption{\label{alg:train_general} The Linear \general MI Attack}
\end{algorithm}

In order to learn the weights of $D$, we partition $\tilde S$ into an ``in'' dataset, $\tilde S^1$, and an ``out'' dataset, $\tilde S^0$.
We train a shadow target model, $\check g$, on $\tilde S^1$ and a proxy model, $\tilde g$, on $\tilde S^0$.
We then create a labeled dataset, $T$, where the features are the weights and biases of $\check g$, the weights and biases of $\tilde g$, and $x$; and the labels are 1 for $x$ belonging to $\tilde S^1$ and 0 for $x$ belonging to $\tilde S^0$.
Finally we train to find the parameters to $D$ that minimize the 0-1 loss, $\mathcal{L}$, of the \general attack on $T$.
We can increase the size of $T$ to improve the generalization of the attack by repeating over multiple in/out splits of $\tilde S$.
This procedure is described in Algorithm~\ref{alg:train_general}.

\subsection{Calibrating for Precision}\label{sec:calibrating}

Recall the ``naive'' attack that predicts that an instance, $x$, is a member of the training set if and only if $x$ was classified correctly.
In practice, this naive approach is not a pragmatic attack because, while it will achieve advantage equal to the target model's generalization error (and close to that of prior black-box approaches~\cite{shokri16shadow}), the only way to evaluate the confidence of the inference is to use the target model's own confidence score.
As most neural networks are not well-calibrated~\cite{guo2017calibration}, this makes it difficult to form confident inferences.
On the other hand, the derivation in Section~\ref{sec:bayes-opt-mi} suggests a direct probabilistic interpretation of the attack model's output.
While the \emph{maximum likelihood estimator}, which predicts $x$ is a member of the training set when $\p{T\ |\ X=x, Y=y} > \half$, maximizes accuracy, the precision, and therefore confidence in positive inferences, is increased by increasing the decision threshold above $\half$.

Under the Gaussian Naive Bayes assumption, the probability given by $m$ is exact, and there is no issue with calibration by this approach.
As a matter of practice, there are two main concerns.
First, the training set is finite, so the recall will drop to zero at some point as the threshold is raised for greater precision.
Second, if the generative assumptions are violated, the confidence may not correspond to an exact probability.
We must therefore be careful when selecting a decision threshold.

Calibrating the decision threshold for the desired precision/recall trade-off requires access to the training set, $S$. 
However, the attack model is obtained using $\tilde S$, which is disjoint from $S$.
Instead, we can stipulate that \emph{the elements of $\tilde S$ are to be classified as non-members} for the purpose of calibration, and use the following heuristic:
given a false-positive tolerance parameter $\alpha$, set the threshold $\tau_y$ for each class $y$ as the $\alpha^{th}$-percentile confidence score of a sample of $\tilde S$ belonging to class $y$.
This is detailed in Algorithm~\ref{alg:calibrate}.
In Section~\ref{sec:eval_precision}, we show that this heuristic consistently increases the precision of our attack on real data.

\begin{algorithm}[t] 
\small
\Fn{\FcalibrateThreshold($m$, $\tilde S$, $\alpha$)}{
    $\tilde S'\ \leftarrow\ \Fsample(\tilde S)$\;
    $\tilde P'_y\ \leftarrow\ 
        [m^{y'}(x')\ \text{for}\ (x',y')\in\tilde S' : y' = y]
        \quad \forall y\in[C]$\;
    $\tau_y\ \leftarrow\ \Fsort(\tilde P'_y)_{\alpha|\tilde P'_y|}
        \quad \forall y\in[C]$\;
    \KwRet $\tau$\;
}
\Fn{\Fpredict($m$, $x$, $y$, $\tau$)}{
    \KwRet $1$ if $m^y(x) > \tau_y$ else $0$\;%
}
\caption{\label{alg:calibrate} Calibrating the Decision Threshold}
\end{algorithm}

\section{Membership Inference in Deep Models}\label{sec:deep}

We showed how to approximate the Bayes-optimal estimator for membership prediction using the weights of a linear target and proxy model in Section~\ref{sec:categorical}.
In this section, we extend the same reasoning to deep models.
However, as deep networks learn novel intermediate representations, the \emph{semantic meaning} of an internal feature at a given index---i.e., the data characteristic that it associates with---will not necessarily line up with the semantic meaning of the corresponding internal feature in another model~\cite{Bengio2012,YosinskiCBL14}.
This holds even when the models share identical architectures, training data, and hyper-parameters, as long as the randomization in the gradient descent is unique.
In general, the only features for which two models will necessarily agree are the models' inputs and outputs, as these are not defined by the training process.

This poses a challenge for any white-box attack that attempts to extend the ``shadow model'' approach~\cite{shokri16shadow} developed for black-box membership inference.
Consider such an approach, which learns properties of internal features that indicate membership---involving activations, gradients, or any other quantity---from shadow models.
Any such property must make reference to specific internal features within the shadow model, but even if the target model contains internal features that match these properties, they are unlikely to reside at exactly the same location within the network as they do in the shadow model.
\emph{This is why previous white-box attacks~\cite{nasr18whitebox} require large amounts of the target model's training data}; rather than learning attack models from shadow models, they are forced to learn them from the target model itself and its training data.

To circumvent this limitation, one must either construct a mapping between internal features in the shadow and target models, or fix the feature representation in the shadow model to preserve semantic meaning between the two.
In this section, we show how to accomplish the latter by constructing a series of \emph{local linear approximations} of the network (Section~\ref{sec:linear_approx}), one for each internal layer, that operate on the feature representation of the target model.
Because each approximation is linear, we can apply any of the attacks from Section~\ref{sec:analysis} to each approximation, and combine the results (Section~\ref{sec:combining_layers}) to form an attack model for the full network.

\subsection{Local Linear Approximations of Deep Models}\label{sec:linear_approx}

\begin{algorithm}[t] 
\small
\Fn{\FmakeModel($\hat g \circ \hat h$, $\tilde S$)}{
	$\tilde S'\ \leftarrow\ [(\hat h(x), y)\ \text{for}\ (x,y) \in \tilde S]$\;
    $\tilde g\ \leftarrow\ \FtrainProxy(\tilde S')$\;
    $w^y\ \leftarrow\ \lambda(z) : 
    	\chi(\hat g\circ\hat h, P^z_0)_y - 
    	\chi(\tilde g\circ\hat h, P^z_0)_y
    	\quad \forall y\in[C]$\;
    $b^y\ \leftarrow\ \hat g(0)_y - \tilde g(0)_y\quad \forall y\in[C]$\;
    \KwRet $\lambda (x,y) : \sigmoid{{w^y(\hat h (x))}^T \hat h(x) + b^y}$\;
}
\Fn{\Fpredict($m$, $x$, $y$)}{
    \KwRet $1$ if $m^y(x) > \half$ else $0$\;
}
\caption{\label{alg:bayes_deep} The Deep \bayesian MI Attack}
\end{algorithm}

We define a local linear approximation in terms of a \emph{slice}, $\langle g, h\rangle$, which decomposes a deep  network, $f$, into two functions, $g$ and $h$, such that $f = g \circ h$. 
Intuitively, a slice corresponds to a layer, $\ell$, of the network, where $h$ computes the features that are input to layer $\ell$, and $g$ computes the output of the model from these features.

For the slice at the top layer of the network,
$g$ is simply a linear model acting on features computed by the rest of the model.
In this case no local approximation is needed and the \bayesian (Algorithm~\ref{alg:bayes}) and \general (Algorithm~\ref{alg:train_general}) attacks can by applied directly to $g$ using internal features that are precomputed by $h$.

For slices lower in the network, $g$ is no longer linear, but we can approximate the way in which $g$ makes use of its features at a particular point by constructing a linear model that agrees with it \emph{at that point}.
To do this, we make use of an \emph{influence measure} over the inputs of $g$ to its computed output for each point.
Given a model, $f$, a point, $x$, and feature, $j$, the \emph{influence} $\chi_j(f, x)$ of $x_j$ on $f$ is a quantitative measure of $x_j$'s contribution to the output of $f$.
A growing body of work on influence measures~\cite{leino18influence,simonyan13saliency,sundararajan17integrated} provides several choices for $\chi$, each with different properties.

For this approximation, we propose using an influence measure that \emph{(1)} works on internal features, \emph{(2)} weights features according to their individual marginal contribution to the model's output, \emph{(3)} satisfies \emph{linear agreement}, and \emph{(4)} is \emph{efficient with respect to a chosen baseline}.
Linear agreement requires that when $f$ is linear, the influence of feature $x_j$ is simply the corresponding weight, $W_j$.
Thus, the influence measure generalizes the notion of weights in a linear model, and we can use the influence of a feature in place of the corresponding weight in Equation~\ref{eq:opt_w_b}, while obtaining the same result. 
However, in order for this substitution to work at a particular internal point, $z = h(x)$, we also require that $g(z) = \bar W_x^T z + \bar b$, where $\bar W_x$ captures how each of the features, $z_j$, are used to obtain the model's output, which is semantically meaningful, at point, $x$.
This follows if $\chi$ is \emph{efficient} with respect to a \emph{baseline} point $z^0$, as defined in Equation~\ref{eq:completeness}.
\begin{equation}\label{eq:completeness}
	\sum_{j}{\chi_{j}(g\circ h, z)(z_j - z^0_j)} = g(z) - g(z^0)
\end{equation}
When (\ref{eq:completeness}) holds, we can set $z^0$ to zero to arrive at the desired local linear approximation, noting that efficiency with respect to the zero baseline implies $g(z) = \chi(g \circ h, z)^T z + g(0)$.

The unique influence measure satisfying the first three properties is \emph{internal influence}~\cite{leino18influence}, given by Equation~\ref{eq:infl_alt}. 
Note that rather than operating on a single point, this measure operates over a \emph{distribution of interest}, $P$, which specifies a distribution of points in the model's latent space, $z = h(x)$.
\begin{equation}\label{eq:infl_alt}
	\chi_j(g\circ h, P) = \int\displaylimits_{z\in h(\mathcal X)}{
		\frac{\partial g}{\partial z_j}\Bigg\vert_{z}P(z)dz}
\end{equation}
When we set $P$ to the uniform distribution over the line from a baseline $z^0$ to $z$, denoted $P^{z}_{z_0}$, then this measure also satisfies efficiency in exactly the manner described above.
We can therefore locally approximate $g$ at $z$ as $\bar{g}(z) = \bar W_x^T z + \bar b$, where $\bar W_x = \chi(g\circ h, P^{z}_{0})$ and $b = g(0)$. 

Thus, we can apply the attacks in Algorithm~\ref{alg:bayes} and Algorithm~\ref{alg:train_general} (Section~\ref{sec:analysis}) on an arbitrary layer of a deep network, by locally approximating the remainder of the network as a linear model at each point the attack is applied to.
Note that this gives a separate set of weights for each input, $x$ (hence why we call the approximation ``local''); however, our attacks are parametric in the weights of the target model, so only a single attack model is necessary.
The modification of Algorithm~\ref{alg:bayes} for an arbitrary slice, $\langle\hat g, \hat h\rangle$, of a target deep network, $\hat f$, is detailed in Algorithm~\ref{alg:bayes_deep}.
An analogous modification of Algorithm~\ref{alg:train_general} follows as well, by simply replacing each reference to weights with influence measurements, but is omitted for the sake of brevity.

\paragraph{Summary}
We can generalize the attacks given by Algorithms \ref{alg:bayes} and \ref{alg:train_general} to apply to an arbitrary layer of a deep target network by replacing the weights with their natural generalization, \emph{influence}.
Because influence allows us to create a faithful local linear approximation of the model for any given point, this generalized attack follows from the same analysis on linear models from Section~\ref{sec:analysis}.
In Section~\ref{sec:combining_layers}, we suggest a method for combining attacks on each individual layer to create an attack that utilizes white-box information from all the layers of a deep network.

\subsection{Combining Layers}\label{sec:combining_layers}

The results of Section~\ref{sec:linear_approx} allow us to leverage overfitting in each learned representation employed by the target model towards membership inference.
Attacks on different layers may pick up on different signals, but because the model's internal representations are not independent across layers, we cannot simply concatenate the approximated weights of each layer and treat it as an attack on a single model.
Instead, we make use of a \emph{meta model}, which learns how to combine the logistic outputs of the individual layer-wise attacks.
The meta model takes the confidences of the attack defined in Section~\ref{sec:linear_approx} applied to each layer, and outputs a single decision.

To train a meta model, $m'$, to attack target model, $f$, we partition $\tilde S$ into two parts, $\tilde S^1$ and $\tilde S^0$.
We train a shadow target model, $\check f$, on $\tilde S^1$.
Then, for each layer, $\ell$, in $f$, we train an attack model, $m_\ell$, on the $\ell^{th}$ layer of $\check f$, as described in Section~\ref{sec:linear_approx}.
We then construct a training set, $T = T^1 \cup T^0$, such that $(x', y')\in T^1$ is constructed as $(x'_\ell, y') = (m_\ell^y(x), 1)$ for $(x, y)\in \tilde S^1$, and $(x', y') \in T^0$ is constructed as $(x'_\ell, y') = (m_\ell^y(x), 0)$ for $(x, y)\in \tilde S^0$.
We can increase the size of $T$ by creating multiple random partitions of $\tilde S$.
Finally, we train $m'$ on $T$.

When building a meta model for the \general attack, we can train $m'$ jointly with the displacement metric, $d$, rather than first learning a \general attack on each layer.
We also use a separate distance metric, $d_\ell$ for each layer, $\ell$, of $f$.

\section{Evaluation}\label{sec:eval}

In this section, we aim to answer several questions about the attacks described in Sections \ref{sec:analysis} and \ref{sec:deep} using empirical results on several real and synthetic datasets.
Section~\ref{sec:defenses} presents additional experimental results having to do with the efficacy of several popular defenses against our attacks.


\emph{\textbf{How sensitive are our attacks to the data assumptions made in Section~\ref{sec:analysis}, hyperparameter choices, and amount of data?}}
In Section~\ref{sec:eval_omniscient}, we find 
that the learning-based attack described in Section~\ref{sec:generalizing} (\general) recovers nearly all of the advantage of the optimal ``omniscient'' attack, despite making no generative assumptions.
Additionally, we show how the hyperparameters used in this attack can be effectively tuned using validation data.
Finally, Section~\ref{sec:eval_data_scaling} discusses attack performance as more or less data is available both for training and to the attacker.

\emph{\textbf{Do certain layers leak more training information than others?}}
Section~\ref{sec:eval_layers} explores the effectiveness of the meta attack model described in Section~\ref{sec:combining_layers} at combining predictions from attacks on each layer of the model. 
Our results show that while all layers play a role in leaking information, in some cases attacks which use combined information from different layers have greater efficacy than the corresponding sum of layer-wise independent attacks.

\emph{\textbf{Relative to prior attacks on real data: (1) are the \bayesian and \general attacks more effective in terms of overall accuracy? (2) does the calibration step (Section~\ref{sec:calibrating}) consistently lead to more confident inferences? (3) do our attacks work on well-generalized models?}}
Our results in Section~\ref{sec:eval_comparison} indicate that \bayesian and \general improve on the performance of prior black-box attacks, both in terms of accuracy and to a larger extent precision.
Moreover, even on models low generalization error ($<2\%$), our attack can be calibrated make high-confidence inferences, which we find is not possible with prior approaches.

\subsection{Experimental Setup}\label{sec:exp_setup}

We now present details on the datasets, target models, methodology, and attack methods used in our experiments.

\paragraph{Datasets} 
We performed experiments over both synthetic data and nine classification datasets derived from real data.
In general, we chose datasets from domains, such as medicine and finance, for which membership inference is likely to be a real concern.
To facilitate a baseline for comparison against prior work, we also included three common image datasets (MNIST, CIFAR10, and CIFAR100) that are less-plausibly connected to privacy, but serve as effective benchmarks, particularly because they have been studied in nearly all published membership inference experiments.

The synthetic data were generated with 10 classes, 75 features, and 400, 800, or 1,600, records, with an equal number of records per class. 
The features, $x_j$, of the synthetic data were drawn randomly from a multivariate Gaussian distribution with parameters, $\mu_{y}$ (for each class, $y$) and $\Sigma$, where $\mu_{yj}$ was drawn uniformly at random from $[0,1]$, and $\Sigma$ was a diagonal matrix with $\Sigma_{jj}$ drawn uniformly at random from $[0.5,1.5]$.

Among the classification datasets were \emph{Adult}, \emph{Pima Diabetes} (obtained from the UCI Machine Learning Repository); \emph{Breast Cancer Wisconsin}, \emph{Hepatitis}, \emph{German Credit}, \emph{Labeled Faces in the Wild} (obtained from scikit-learn's \texttt{datasets} API); \emph{MNIST}~\cite{mnist}, \emph{CIFAR10}, and \emph{CIFAR100}~\cite{cifar}.
Figure~\ref{fig:train_test} shows the characteristics of each of these datasets.


\paragraph{Target Models}
The target models we used to conduct our experiments include linear models, multi-layer perceptrons, and convolutional neural networks.
Each model was trained until convergence with categorical cross-entropy loss, using SGD with a learning rate of $0.1$, a decay rate of $10^{-4}$, and Nesterov momentum. 

Linear models were implemented as a single-layer network in Keras~\cite{keras} using a softmax activation.
We used linear models only for the synthetic data.
For non-image real data, we used a multi-layer perceptron (MLP) with one hidden layer and $ReLU$ non-linearities, implemented in Keras.
For datasets with $n$ features, we employed $2n$ hidden units, followed by a softmax layer with one unit per class.
For image data, we used a CNN architecture based on LeNet, with two convolutional layers with $5\times 5$ filters and $20$ and $50$ output channels respectively (each convolutional layer is followed by a max pooling layer), followed by a fully connected layer with $500$ neurons.
We trained CNNs with a 25\% dropout rate following each pooling layer, and a 50\% dropout rate following the fully connected layer.


Each target model is a pair containing an architecture and a dataset. 
We refer to each target model by its dataset abbreviation given in Figure~\ref{fig:train_test}.
The train and test accuracy for each of the target models used in our evaluation are given in the final two columns of Figure~\ref{fig:train_test}.

\begin{figure}
	\input{figures/model_train_test}
	\caption{\label{fig:train_test}
		Characteristics of the datasets and models used in our experiments.
	}
	\vspace{-.5em}
\end{figure}

\paragraph{Methodology}
When evaluating each attack, we randomly split the data into three disjoint groups: \emph{train}, \emph{test}, and \emph{hold-out}.
The train and test groups were each comprised of one fourth of the total number of instances, and the hold-out group contained the remaining one half of the instances.
The target model was trained on the train group, while the attacks were allowed to make use of the hold-out group only.
The attack model's predictions were evaluated on the train group (members) and the test group (non-members).
Each experiment was repeated 10 times over different random samplings of the data split, and the results were averaged.

\paragraph{Attack Methods}
Throughout our evaluation, we assess four different attacks: \naive, \bayesian, \general, and \bb.
The \naive attack refers to the simple attack introduced in Section~\ref{sec:intro}, in which the attack model predicts an instance, $x$, is a member of the training set if and only if $x$ was classified correctly.

For the \bayesian attack (introduced in Section \ref{sec:categorical}), we trained 10 proxy models on random samples from the hold-out group, and took the mean of their approximated weights at each point for added robustness.
When attacking MLP models, we performed the attack on the final layer of the MLP using Algorithm~\ref{alg:bayes}.
When attacking LeNet models, we used a meta attack model (described in Section~\ref{sec:combining_layers}) that was trained on data from 10 shadow models trained on 10 samples from the hold-out group.
We used a MLP with 16 internal neurons for the meta model and trained it for 32 epochs with Adam~\cite{kingma15}.

For the \general attack (introduced in Section~\ref{sec:generalizing}), we construct an attack model that learns a displacement function, $D_\ell$ (Algorithm~\ref{alg:train_general}), for each layer, $\ell$, of the network, and combines the results with a meta attack model, $M$.
The attack model was trained for 32 epochs with Adam, using data from 10 shadow models trained on the hold-out group.
As suggested in Section~\ref{sec:generalizing}, we modeled each $D_\ell$ as a convolutional neural network.
In each experiment, the networks modeling $M$ and each $D_\ell$ had at most one hidden layer, with $n_M$ and $n_D$ hidden units, respectively (in our experiments each $D_\ell$ used the same architecture, though this need not be the case in general).
In order to determine $n_M$ and $n_D$ for each dataset, we created a validation set using 10 shadow models trained on different random splits of the hold-out group, and performed a parameter sweep over $n_M, n_D$.
We then took the $n_M$ and $n_D$ yielding the highest validation accuracy for each target model.
We find that because the attack model is highly regularized via its restrictive architecture, the validation accuracy is a reasonably good indicator of the test accuracy, making it a useful tool for hyper-parameter tuning (see Figure~\ref{fig:validation}).

The \bb attack refers to the black-box shadow model attack~\cite{shokri16shadow}, explained briefly in Section~\ref{sec:related}.
In each experiment, the \bb attack was trained using 10 shadow models trained on 10 samples from the hold-out group.


\subsection{Sensitivity to Assumptions \& Hyper-parameters}\label{sec:eval_omniscient}

In Section~\ref{sec:bayes-opt-mi}, we derive the Bayes-optimal membership inference attack on Gaussian data satisfying the naive-Bayes condition.
The weights of the optimal membership predictor for this case, given by Theorem~\ref{thm:opt_wb}, are a function of the empirical training distribution parameters and true distribution of the data, which, of course, would be unknown to an attacker.
Section~\ref{sec:categorical} describes how to address this, using a \emph{proxy model} to capture the difference between the data used to train the target model and the general population.

\begin{figure}[t]
	\input{figures/omniscient_comp}
	\caption{\label{fig:proxy}
		Comparison of the \bayesian and \general attacks to an \textit{omniscient} attack, which has knowledge of $\hat\mu$, $\mu^*$, and $\sigma$, and thus can use Theorem~\ref{thm:bayes-opt} directly without the use of a proxy model. 
		In one case, the \general attack was given the minimum capacity to reproduce the \bayesian attack, i.e., $d$ is simply a weighted sum of $\hat W_i$ and $\tilde W_i$. 
		In another case, the \general attack was given excess capacity, with $16$ hidden units in $d$.
		Three target models, trained on synthetic Gaussian naive-Bayes data with training set sizes of $100$, $200$, and $400$, were attacked.
	}
\end{figure}

Figure~\ref{fig:proxy} demonstrates the effectiveness of the proxy model in our attack, by comparing our \bayesian attack using a proxy model to an ``omniscient'' attack, which uses Equation~\ref{eq:bayes_opt_weights} directly, with knowledge of the train and general distribution.
We can consider the omniscient attack as giving an upper bound on the expected accuracy of a white-box attack on Gaussian naive-Bayes data, as it is the true Bayes-optimal attack (while \bayesian is the approximate Bayes-optimal attack according to Proposition~\ref{thm:opt_wb}).
Our attack achieves on average $84\%$ of the advantage of the omniscient attack, suggesting that the proxy model was able to approximately capture the general distribution as necessary for the purpose of detecting the target model's idiosyncratic use of features.

In Section~\ref{sec:generalizing}, we further generalize the \bayesian attack to use a learned displacement function that may be more appropriate for distributions that don't resemble the Gaussian naive-Bayes assumption.
While we find that this \general attack often generalizes to arbitrary distributions better than the \bayesian attack, because its displacement function is learned, it is possible for the \general attack to overfit.

Figure~\ref{fig:proxy} also shows the accuracy of the \general attack on Gaussian naive-Bayes data.
When the neural network representing the displacement function is given exactly enough capacity to reproduce the \bayesian attack, \general recovers on average $94\%$ of the advantage of the \bayesian attack.
Upon inspecting the weights of the displacement network, we find that \general learns almost exactly element-wise subtraction, demonstrating its potential to learn the optimal displacement function.
When given excess capacity, the \general attack performs only marginally worse, achieving on average $92\%$ of the minimal \general attack's advantage ($86\%$ of \bayesian), suggesting that \general is not highly prone to overfitting.

\begin{figure}[t]


\pgfplotstableread[row sep=\\,col sep=&]{
  X    & validation & test \\
  a00  & 62.7       & 58.2 \\
  a80  & 59.0       & 57.0 \\
  a160 & 64.0       & 61.8 \\
  a04  & 59.6       & 56.4 \\
  a84  & 59.7       & 58.7 \\
  a164 & 63.4       & 60.0 \\
  a08  & 62.4       & 60.5 \\
  a88  & 60.4       & 58.9 \\
  a168 & 63.4       & 58.3 \\
}\HepVal


\centering
\resizebox{\linewidth}{!}{%
\tikzstyle{every node}=[font=\small]
\begin{tikzpicture}
  \begin{groupplot}[
      group style={
        group name=myplot,
        group size= 1 by 1,
        ylabels at=edge left,
        yticklabels at=edge left,
        horizontal sep=0pt,
        vertical sep=16pt},
      footnotesize,
      width=0.99\columnwidth,
      scale only axis,
      height=.25\columnwidth,
      tickpos=left,
      ytick align=inside,
      ylabel shift = -4pt,
      y tick label style={anchor=east},
      x tick label style={rotate=30, anchor=north east, yshift=5pt, xshift=2pt},
      xtick align=outside,
      ymin=50, ymax=69,
      symbolic x coords={a00, a80, a160, a04, a84, a164, a08, a88, a168},
      xtick=data,
      xticklabels={{(0,0)}, {(8,0)}, {(16,0)}, {(0,4)}, {(8,4)}, {(16,4)}, {(0,8)}, {(8,8)}, {(16,8)}},
      enlarge x limits=0.1,
      enlarge y limits=false,
      extra y ticks = 50,
      extra y tick labels = ,
      extra y tick style = {grid = major},
    ]
    \nextgroupplot[ylabel={\textit{accuracy}}]
        \coordinate (top) at (rel axis cs:0.204,1.0);
        \addplot[
            mark=triangle*,
            dotted,
            mark options={solid, fill=lightgray},
            draw=black,
          ] table[y=validation,x=X]{\HepVal};\label{plots:val}
        \addplot[
            mark=*,
            dotted,
            mark options={solid, fill=darkgray},
            draw=black,
          ] table[y=test,x=X]{\HepVal};\label{plots:tst}

        \end{groupplot}
  \node[draw,yshift=-.8em,xshift=14.1em] at (top) {
      \ref{plots:val} validation
      \ref{plots:tst} test};
\end{tikzpicture}}
	\vspace{-1.5em}
	\caption{\label{fig:validation}
		Plot showing the validation (known to the attacker) and test (unknown to the attacker) accuracies of the \general attack for various attack model architectures on the Hepatitis dataset.
		Each architecture, listed on the x-axis, is represented by a pair, $(n_D, n_M)$, where $n_D$ and $n_M$ are the number of hidden units in the \emph{distance function network} and \emph{meta model network} respectively (see Section~\ref{sec:exp_setup}).
	}
\end{figure}
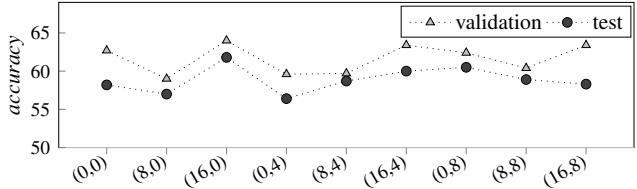

\paragraph{Tuning the \general Attack}
As mentioned, even an over-parameterized displacement function may be able to perform nearly optimally on models trained on simple datasets, like the Synthetic dataset.
However, as the \general attack involves several hyper-parameters, it may be useful to tune these parameters in a reliable way.
We note that an arbitrary number of shadow models can be produced by sampling from the hold-out data, allowing us to construct a validation set on which to evaluate various architectures for implementing the distance function, $D_\ell$, and meta model, $M$, comprising the \general attack.
Figure~\ref{fig:validation} shows an example of the validation accuracy obtained using various architectures for $D_\ell$ and $M$, along with the corresponding test accuracy (unknown to the attacker).
We see that the test accuracy fairly closely follows the validation accuracy, with the maximum for both metrics occurring for the same architecture.
This suggests that the validation accuracy is a reasonably good indicator of the test accuracy making it a useful tool for hyper-parameter tuning.
This is perhaps not too surprising, as the attack model is highly regularized via its restrictive architecture.

\subsection{Data Scaling}\label{sec:eval_data_scaling}

The ``omniscient'' attack developed in Section~\ref{sec:bayes-opt-mi} relies on measuring a difference between the parameters of the \emph{true} data-generating distribution, $\distrib^*$ and the \emph{empirical} distribution, $\hat\distrib$.
Because $\hat\distrib$ is derived from a sample drawn from $\distrib^*$, in expectation $\hat\distrib = \distrib^*$;
that is, as the number of samples in the training set goes to infinity, the true and empirical distributions will converge, rendering even the optimal attack ineffective (0 advantage).
We would therefore expect that for a sufficiently large training set, the success of any MI attack would decline.
Conversely, we may expect the opportunity for better MI performance for smaller training sets.
Indeed, in accordance with this observation, we see that even the omniscient attack sees accuracy inversely proportional to the dataset size (Figure~\ref{fig:proxy}).

\begin{figure}[t]


\pgfplotstableread[row sep=\\,col sep=&]{
  X      & bayes & general \\
  0.125  & 0.521 & 0.535 \\
  0.25   & 0.516 & 0.526 \\
  0.5    & 0.512 & 0.521 \\
  1.0    & 0.508 & 0.516 \\
}\AdultDatascaling

\centering
\resizebox{\linewidth}{!}{%
\tikzstyle{every node}=[font=\small]
\begin{tikzpicture}
  \begin{groupplot}[
      group style={
        group name=myplot,
        group size= 1 by 1,
        ylabels at=edge left,
        yticklabels at=edge left,
        horizontal sep=0pt,
        vertical sep=16pt},
      footnotesize,
      width=0.99\columnwidth,
      scale only axis,
      height=.25\columnwidth,
      tickpos=left,
      ytick align=inside,
      ylabel shift = -4pt,
      y tick label style={anchor=east},
      xtick align=outside,
      ymin=.495, ymax=.545,
      xtick=data,
      xticklabels={$1/8$, $1/4$, $1/2$, $1$},
      enlarge x limits=0.1,
      enlarge y limits=false,
      extra y ticks = .50,
      extra y tick labels = ,
      extra y tick style = {grid = major},
    ]
    \nextgroupplot[ylabel={\textit{accuracy}},xlabel={\textit{fraction of data used}}]
        \coordinate (top) at (rel axis cs:0.204,1.0);
        \addplot[
            mark=triangle*,
            dotted,
            mark options={solid, fill=lightgray},
            draw=black,
          ] table[y=general,x=X]{\AdultDatascaling};
        \addplot[
            mark=*,
            dotted,
            mark options={solid, fill=lightgray},
            draw=black,
          ] table[y=bayes,x=X]{\AdultDatascaling};
 
        \end{groupplot}
  \node[draw,fill=white,yshift=-.85em,xshift=12.755em] at (top) {
      \ref{plots:pl1} \bayesian
      \ref{plots:pl2} \general
  };
\end{tikzpicture}}
	\vspace{-1.5em}
	\caption{\label{fig:data_scaling}
		Accuracy of the \bayesian and \general attacks on the Adult dataset, as the amount of data is scaled from 6,105 records (1/8 of the full dataset) to 48,841 records.
	}
\end{figure}
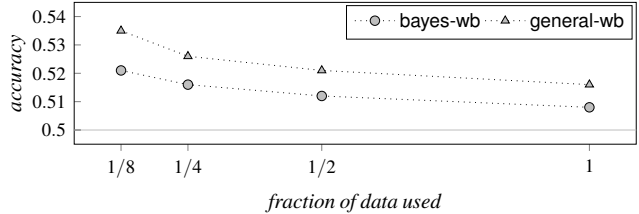

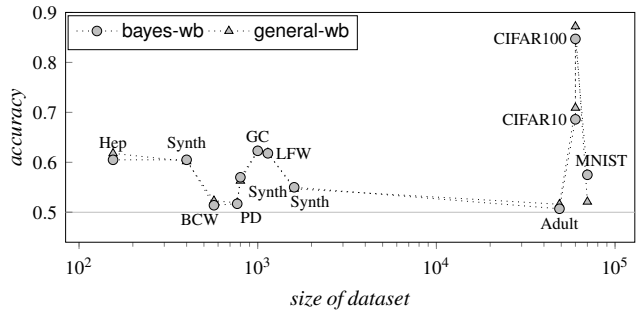
\begin{figure}[t]
	\pgfplotstableread[row sep=\\,col sep=&]{
  X     & bayes & general & dset     & alignment \\
  155   & .605 & .618     & Hep      & -90    \\
  400   & .605 & .602     & Synth    & -90    \\
  569   & .514 & .523     & BCW      & 45\\
  768   & .517 & .519     & PD       & 135 \\
  800   & .570 & .563     & Synth    & 150 \\
  1000  & .623 & .622     & GC       & -90\\
  1140  & .618 & .619     & LFW      & 180\\
  1600  & .550 & .547     & Synth    & 135\\
  48841 & .507 & .516     & Adult    & 90\\
  60000 & .686 & .709     & CIFAR10  & 0\\
  60000 & .847 & .872     & CIFAR100 & 0\\
  70000 & .575 & .521     & MNIST    & -135\\
}\AdultDatascaling

\centering
\resizebox{\linewidth}{!}{%
\tikzstyle{every node}=[font=\small]
\begin{tikzpicture}
  \begin{groupplot}[
      group style={
        group name=myplot,
        group size= 1 by 1,
        ylabels at=edge left,
        yticklabels at=edge left,
        horizontal sep=0pt,
        vertical sep=16pt},
      footnotesize,
      width=0.99\columnwidth,
      scale only axis,
      height=.4\columnwidth,
      tickpos=left,
      ytick align=inside,
      ylabel shift = -4pt,
      y tick label style={anchor=east},
      xtick align=outside,
      ymin=.44, ymax=.9,
      xmode=log,
      xtick={100,1000,10000, 100000},
      enlarge x limits=0.1,
      enlarge y limits=false,
      extra y ticks = .50,
      extra y tick labels = ,
      extra y tick style = {grid = major},
    ]
    \nextgroupplot[ylabel={\textit{accuracy}},xlabel={\textit{size of dataset}}]
        \coordinate (top) at (rel axis cs:0.204,1.0);
        \addplot[
            mark=triangle*,
            dotted,
            mark options={solid, fill=lightgray},
            draw=black,
          ] table[y=general,x=X]{\AdultDatascaling};
        \addplot[
            mark=*,
            dotted,
            mark options={solid, fill=lightgray},
            draw=black,
            visualization depends on={value \thisrow{alignment}\as\algnmnt},
            nodes near coords,
            point meta=explicit symbolic,
            every node near coord/.style={anchor={\algnmnt}, font=\scriptsize}
          ] table[y=bayes,x=X, meta index=3, row sep=\\,col sep=&]{
            X     & bayes & general & dset     & alignment \\
            155   & .605 & .618     & Hep      & -90    \\
            400   & .605 & .602     & Synth    & -90    \\
            569   & .514 & .523     & BCW      & 45\\
            768   & .517 & .519     & PD       & 135 \\
            800   & .570 & .563     & Synth    & 150 \\
            1000  & .623 & .622     & GC       & -90\\
            1140  & .618 & .619     & LFW      & 180\\
            1600  & .550 & .547     & Synth    & 135\\
            48841 & .507 & .516     & Adult    & 90\\
            60000 & .686 & .709     & CIFAR10  & 0\\
            60000 & .847 & .872     & CIFAR100 & 0\\
            70000 & .575 & .521     & MNIST    & -135\\
          };
 
        \end{groupplot}
  \node[draw,fill=white,yshift=-.85em,xshift=1.3em] at (top) {
      \ref{plots:pl1} {\bayesian}
      \ref{plots:pl2} {\general}
  };%

\end{tikzpicture}}
	\vspace{-1.5em}
	\caption{\label{fig:data_scaling_all}
		Accuracy of the \bayesian and \general attacks on each of the datasets in our evaluation, plotted against the size of the respective dataset.
	}
\end{figure}

We find that this pattern persists for real-world datasets as well.
Figure~\ref{fig:data_scaling} shows the accuracy of our attacks on models trained on subsets of various sizes of the Adult dataset (the dataset containing the most records as compared to the number of parameters in the respective model).
We observe that as more data becomes available for training, the advantage of the attack diminishes, becoming quite small ($< 4\%$) on the entire dataset (48,841 records).
This may suggest that the Adult dataset is sufficiently large to preclude any significant information leakage via a modestly-sized MLP model obtained through standard training.

Figure~\ref{fig:data_scaling_all} shows the accuracy of our attacks on each of the datasets used in our evaluation, plotted against the size of the respective dataset.
We see to some extent the same downwards trend as dataset size increases, though there is more noise, and some of the image datasets (especially CIFAR10 and CIFAR100) provide notable exceptions.
This is likely due to the variation in the number of features, the network capacity, and the generalization error across datasets.



\subsection{Combining Layers}\label{sec:eval_layers}

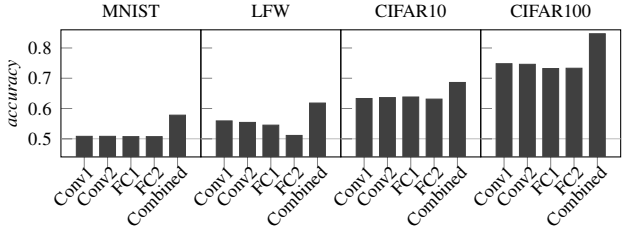
\begin{figure}[t!]
	\resizebox{\columnwidth}{!}{

\centering
\tikzstyle{every node}=[font=\scriptsize]
\begin{tikzpicture}
  \begin{groupplot} [
      group style={
        group name=myplot2,
        group size= 4 by 1,
        xlabels at=edge bottom,
        xticklabels at=edge bottom,
        ylabels at=edge left,
        yticklabels at=edge left,
        horizontal sep=0pt,
        vertical sep=16pt},
      width=0.22\columnwidth,
      scale only axis,
      height=.2\columnwidth,
      tickpos=left,
      ybar,
      ylabel shift = -4pt,
      /pgf/bar width=6pt,
      ytick align=inside,
      y tick label style={anchor=east},
      xtick=data,
      x tick label style={rotate=45, anchor=north east, yshift=4pt, xshift=5pt},
      ymin=.44, ymax=.86,
      enlarge x limits=0.25,
      symbolic x coords={
        Conv1,
        Conv2,
        FC1,
        FC2,
        Combined},
      ytick={.50,.60,.70,.80},
      extra y ticks = .50,
      extra y tick labels = ,
      extra y tick style = {grid = major},
    ]
    \nextgroupplot[
      ylabel={\textit{accuracy}},
      title style={
        at={(0.5,0.9)}, anchor=south}, 
      title=MNIST]
      \addplot [
          ybar, 
          draw=darkgray, 
          fill=darkgray
        ]
        plot
        table {
          x        y   
          Conv1    .508
          Conv2    .508
          FC1      .507
          FC2      .507
          Combined .578
        };
    \nextgroupplot[
      title style={
        at={(0.5,0.9)}, anchor=south}, 
      title=LFW]
      \addplot [
          ybar,
          draw=darkgray, 
          fill=darkgray
        ]
        plot
        table {
          x        y   
          Conv1    .559
          Conv2    .554
          FC1      .545
          FC2      .511
          Combined .618
        };
    \nextgroupplot[
      title style={
        at={(0.5,0.9)}, anchor=south},
      title=CIFAR10]
      \addplot [
          ybar, 
          draw=darkgray, 
          fill=darkgray
        ]
        plot
        table {
          x        y   
          Conv1    .633
          Conv2    .636
          FC1      .638
          FC2      .631
          Combined .686
        };
    \nextgroupplot[
      title style={
        at={(0.5,0.9)}, anchor=south},
      title=CIFAR100]
      \addplot [
          ybar, 
          draw=darkgray, 
          fill=darkgray
        ]
        plot
        table {
          x        y   
          Conv1    .748
          Conv2    .746
          FC1      .732
          FC2      .733
          Combined .847
        };
  \end{groupplot}
\end{tikzpicture}}
	\vspace{-2em}
	\caption{\label{fig:combined_layers}
		Accuracy of the \bayesian attack on each individual layer of LeNet, compared with the accuracy using the combined meta-model. 
	}
\end{figure}

For deep models in particular, we want to be able to use information from each layer in our attack.
In Section~\ref{sec:combining_layers}, we describe a meta attack that combines the outputs of an individual attack on each layer.
Figure~\ref{fig:combined_layers} shows the accuracy of the \bayesian attack on each individual layer and of the meta attack on each LeNet target model.

In every instance, the meta attack is able to substantially outperform any individual attack, indicating that the information it receives from each layer is not entirely redundant.
Moreover, this suggests that \emph{information leakage occurs in the representations learned by layers throughout the model}---that is, each layer plays some role in the leakage of information about the training data.
A possible consequence of this that we hypothesize in Section~\ref{sec:defenses} is that models trained with transfer learning may leak less information about the training data used to tune the model.

Remarkably, for MNIST, the advantage of the meta attack is greater than that of all the individual layers combined.

\subsection{Comparison to Prior Work}\label{sec:eval_comparison}\label{sec:eval_precision}


\begin{figure*}[t!]
	\input{figures/all_methods_apr}
	\caption{\label{fig:all_apr}
		Comparison of the accuracy, precision, and recall of \bayesian and \general with \naive and \bb. 
	}
\end{figure*}

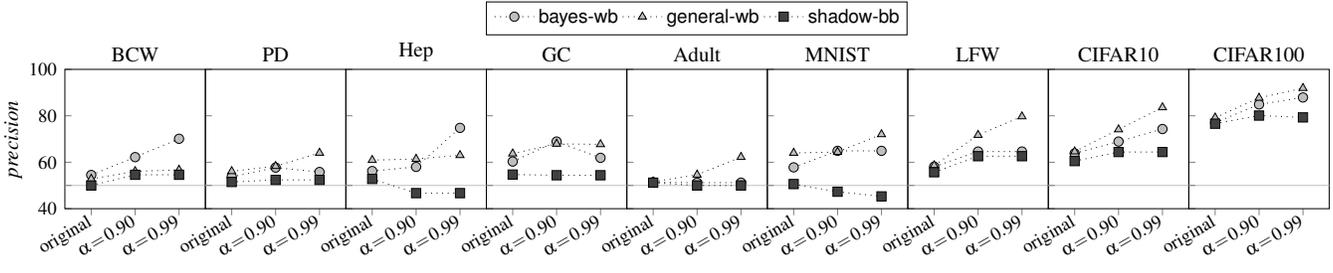
\begin{figure*}[t!]






\pgfplotstableread[row sep=\\,col sep=&]{
  X   & bayesian & general & bb \\
  og  & 54.5     & 52.8    & 50.0 \\
  90t & 62.2     & 56.1    & 54.6 \\
  99t & 70.1     & 56.7    & 54.6 \\
}\BcwRot

\pgfplotstableread[row sep=\\,col sep=&]{
  X   & bayesian & general & bb \\
  og  & 53.7     & 56.1    & 51.5 \\
  90t & 57.7     & 58.3    & 52.4 \\
  99t & 55.8     & 64.0    & 52.4 \\
}\PdRot

\pgfplotstableread[row sep=\\,col sep=&]{
  X   & bayesian & general & bb \\
  og  & 56.2     & 60.9    & 52.8 \\
  90t & 58.0     & 61.4    & 46.7 \\
  99t & 74.8     & 63.0    & 46.7 \\
}\HepRot

\pgfplotstableread[row sep=\\,col sep=&]{
  X   & bayesian & general & bb \\
  og  & 60.3     & 63.7    & 54.7 \\
  90t & 68.9     & 67.9    & 54.4 \\
  99t & 61.9     & 67.7    & 54.4 \\
}\GcRot

\pgfplotstableread[row sep=\\,col sep=&]{
  X   & bayesian & general & bb \\
  og  & 51.4     & 51.6    & 51.2 \\
  90t & 51.2     & 54.6    & 50.0 \\
  99t & 51.2     & 62.2    & 50.0 \\
}\AdultRot

\pgfplotstableread[row sep=\\,col sep=&]{
  X   & bayesian & general & bb \\
  og  & 58.1     & 58.6    & 55.7 \\
  90t & 64.6     & 71.7    & 62.6 \\
  99t & 64.6     & 79.7    & 62.6 \\
}\LfwRot

\pgfplotstableread[row sep=\\,col sep=&]{
  X   & bayesian & general & bb \\
  og  & 57.8     & 64.0    & 50.6 \\
  90t & 64.9     & 64.4    & 47.3 \\
  99t & 64.9     & 72.0    & 45.3 \\
}\MnistRot

\pgfplotstableread[row sep=\\,col sep=&]{
  X   & bayesian & general & bb \\
  og  & 63.8     & 64.6    & 60.5 \\
  90t & 68.9     & 74.1    & 64.4 \\
  99t & 74.3     & 83.6    & 64.4 \\
}\CifarRot

\pgfplotstableread[row sep=\\,col sep=&]{
  X   & bayesian & general & bb \\
  og  & 77.0     & 79.2    & 76.6 \\
  90t & 84.9     & 87.7    & 80.1 \\
  99t & 87.9     & 91.9    & 79.3 \\
}\CifarCRot

\centering
\resizebox{\linewidth}{!}{%
\tikzstyle{every node}=[font=\small]
\begin{tikzpicture}
  \begin{groupplot}[
      group style={
        group name=myplot,
        group size= 9 by 1,
        ylabels at=edge left,
        yticklabels at=edge left,
        horizontal sep=0pt,
        vertical sep=16pt},
      footnotesize,
      width=0.12\textwidth,
      scale only axis,
      height=.25\columnwidth,
      tickpos=left,
      ytick align=inside,
      ylabel shift = -4pt,
      y tick label style={anchor=east},
      x tick label style={rotate=30, anchor=north east, yshift=5pt, xshift=2pt},
      xtick align=outside,
      ymin=40, ymax=100,
      symbolic x coords={og, 90t, 99t},
      xtick=data,
      xticklabels={original, $\alpha = 0.90$, $\alpha = 0.99$},
      enlarge x limits=0.3,
      enlarge y limits=false,
      extra y ticks = 50,
      extra y tick labels = ,
      extra y tick style = {grid = major},
    ]
    \nextgroupplot[title={BCW},ylabel={\textit{precision}}]
        \addplot[
            mark=*,
            dotted,
            mark options={solid, fill=lightgray},
            draw=black,
          ] table[y=bayesian,x=X]{\BcwRot};
        \addplot[
            mark=triangle*,
            dotted,
            mark options={solid, fill=lightgray},
            draw=black,
          ] table[y=general,x=X]{\BcwRot};
        \addplot[
            mark=square*,
            dotted,
            mark options={solid, fill=darkgray},
            draw=black
          ] table[y=bb,x=X]{\BcwRot};
 
    \nextgroupplot[title={PD}]
        \addplot[
            mark=*,
            dotted,
            mark options={solid, fill=lightgray},
            draw=black,
          ] table[y=bayesian,x=X]{\PdRot};\label{plots:pl1}
        \addplot[
            mark=triangle*,
            dotted,
            mark options={solid, fill=lightgray},
            draw=black,
          ] table[y=general,x=X]{\PdRot};\label{plots:pl2}
        \addplot[
            mark=square*,
            dotted,
            mark options={solid, fill=darkgray},
            draw=black
          ] table[y=bb,x=X]{\PdRot};\label{plots:pl3}

    \nextgroupplot[title={Hep}]
        \coordinate (top) at (rel axis cs:0.85,1.2);
        \addplot[
            mark=*,
            dotted,
            mark options={solid, fill=lightgray},
            draw=black,
          ] table[y=bayesian,x=X]{\HepRot};
        \addplot[
            mark=triangle*,
            dotted,
            mark options={solid, fill=lightgray},
            draw=black,
          ] table[y=general,x=X]{\HepRot};
        \addplot[
            mark=square*,
            dotted,
            mark options={solid, fill=darkgray},
            draw=black
          ] table[y=bb,x=X]{\HepRot};

    \nextgroupplot[title={GC}]
        \addplot[
            mark=*,
            dotted,
            mark options={solid, fill=lightgray},
            draw=black,
          ] table[y=bayesian,x=X]{\GcRot};
        \addplot[
            mark=triangle*,
            dotted,
            mark options={solid, fill=lightgray},
            draw=black,
          ] table[y=general,x=X]{\GcRot};
        \addplot[
            mark=square*,
            dotted,
            mark options={solid, fill=darkgray},
            draw=black
          ] table[y=bb,x=X]{\GcRot};

    \nextgroupplot[title={Adult}]
        \addplot[
            mark=*,
            dotted,
            mark options={solid, fill=lightgray},
            draw=black,
          ] table[y=bayesian,x=X]{\AdultRot};
        \addplot[
            mark=triangle*,
            dotted,
            mark options={solid, fill=lightgray},
            draw=black,
          ] table[y=general,x=X]{\AdultRot};
        \addplot[
            mark=square*,
            dotted,
            mark options={solid, fill=darkgray},
            draw=black
          ] table[y=bb,x=X]{\AdultRot};

    \nextgroupplot[title={MNIST}]
        \addplot[
            mark=*,
            dotted,
            mark options={solid, fill=lightgray},
            draw=black,
          ] table[y=bayesian,x=X]{\MnistRot};
        \addplot[
            mark=triangle*,
            dotted,
            mark options={solid, fill=lightgray},
            draw=black,
          ] table[y=general,x=X]{\MnistRot};
        \addplot[
            mark=square*,
            dotted,
            mark options={solid, fill=darkgray},
            draw=black
          ] table[y=bb,x=X]{\MnistRot};

    \nextgroupplot[title={LFW}]
        \addplot[
            mark=*,
            dotted,
            mark options={solid, fill=lightgray},
            draw=black,
          ] table[y=bayesian,x=X]{\LfwRot};
        \addplot[
            mark=triangle*,
            dotted,
            mark options={solid, fill=lightgray},
            draw=black,
          ] table[y=general,x=X]{\LfwRot};
        \addplot[
            mark=square*,
            dotted,
            mark options={solid, fill=darkgray},
            draw=black
          ] table[y=bb,x=X]{\LfwRot};

    \nextgroupplot[title={CIFAR10}]
        \addplot[
            mark=*,
            dotted,
            mark options={solid, fill=lightgray},
            draw=black,
          ] table[y=bayesian,x=X]{\CifarRot};
        \addplot[
            mark=triangle*,
            dotted,
            mark options={solid, fill=lightgray},
            draw=black,
          ] table[y=general,x=X]{\CifarRot};
        \addplot[
            mark=square*,
            dotted,
            mark options={solid, fill=darkgray},
            draw=black
          ] table[y=bb,x=X]{\CifarRot};

    \nextgroupplot[title={CIFAR100}]
        \addplot[
            mark=*,
            dotted,
            mark options={solid, fill=lightgray},
            draw=black,
          ] table[y=bayesian,x=X]{\CifarCRot};
        \addplot[
            mark=triangle*,
            dotted,
            mark options={solid, fill=lightgray},
            draw=black,
          ] table[y=general,x=X]{\CifarCRot};
        \addplot[
            mark=square*,
            dotted,
            mark options={solid, fill=darkgray},
            draw=black
          ] table[y=bb,x=X]{\CifarCRot};

        \end{groupplot}
  \node[draw,yshift=1em,xshift=10em] at (top) {
      \ref{plots:pl1} \bayesian
      \ref{plots:pl2} \general 
      \ref{plots:pl3} \bb};
\end{tikzpicture}}
	\vspace{-1.5em}
	\caption{\label{fig:rule_of_thumb}
		Precision of the \bayesian, \general, and \bb attacks, calibrated using the heuristic outlined described in Algorithm~\ref{alg:calibrate} (with $\alpha = 0.90$ and $\alpha = 0.99$), compared to the precision with no calibration (default threshold).
	}
\end{figure*}

Finally, we compare our approach to previous work, namely, \bb~\cite{shokri16shadow}.
In particular, we compare \emph{(1)} performance in terms of accuracy, precision, and recall; and \emph{(2)} the reliability of the attack confidence when used to calibrate for higher precision.
In short, our results show that both \bayesian and \general outperform \bb, and can be more reliably calibrated to achieve confident inferences for the attacker.
Furthermore, even on some well-generalized models, on which \bb and \naive fare poorly, our attacks can be calibrated to make confident inferences, and sometimes also achieve non-trivial advantage.
Finally, we find that there is often little advantage to \bb over \naive, both because \bb often performs comparably to \naive, and because \bb does not always produce calibrated confidence scores.

\paragraph{Performance}
Figure~\ref{fig:all_apr} shows the accuracy, precision, and recall of \naive, \bayesian, \general, and \bb.
The precision shown is before calibration attack (calibration results are shown in Figure~\ref{fig:rule_of_thumb}).
We see that both \bayesian and \general are consistently more accurate and precise than \naive and \bb. 
At least one of \bayesian or \general obtains the highest accuracy of the four methods on each target except Adult, and \emph{both} outperform the other two methods in terms of precision in all cases.
In some cases, the improvement in accuracy of at least one of our attacks over prior work is by as much as seven percentage points, though in others our accuracy is only modestly better;
however, in terms of precision, the difference is more pronounced in almost every case (typically greater by at least five percentage points).

Typically \naive or \bb achieve the highest recall, but we note that both methods do so with lower precision; and at least in the case of \naive, this is merely a consequence of the fact that most of the models have a high training accuracy.


Our results for the performance of \bb are roughly in line with previously reported results for \bb on the datasets which have been used for evaluation in prior work (Adult, MNIST, LFW, CIFAR10, and CIFAR100)~\cite{shokri16shadow,ndss19salem}.
On CIFAR10 and CIFAR100, our results are slightly lower than the results reported for \bb by \citeauthor{shokri16shadow}, however, our target models trained on CIFAR10 and CIFAR100 use dropout and have a lower generalization error than the models in the attacks reported by \citeauthor{shokri16shadow}, which most likely accounts for this small discrepancy.

\paragraph{Calibration}


As argued in Sections~\ref{sec:intro} and~\ref{sec:background}, one of the key desiderata of a membership inference attack is precision.
In order to calibrate an attack for precision, the confidence outputted by the attack must be informative.
Here, we examine the calibration of the confidence outputs of our attacks compared to \bb\ (\naive does not provide a confidence score with which to calibrate).



We find that increasing the decision threshold of the \bayesian and \general attacks has a positive effect on precision.
In particular, using the heuristic defined in Algorithm~\ref{alg:calibrate}, we are able to consistently improve the precision of our attacks.
Figure~\ref{fig:rule_of_thumb} shows the precision of our attack as the decision threshold is raised according to Algorithm~\ref{alg:calibrate}, for $\alpha = 0.90$, and $\alpha = 0.99$, compared to the uncalibrated attack.
In each case the precision increases, often by 10 or more percentage points.
Though in practice, an attacker would not be easily able to tune the calibration hyper-parameter, $\alpha$, the consistency of the results in Figure~\ref{fig:rule_of_thumb} suggest that values of $0.90$ and $0.99$ serve as a practical ``rule-of-thumb'' for reliable calibration.

On all convolutional models, \general is able to be calibrated to upwards of $75\%$ precision.
Notably, this includes the model trained on MNIST, which has only $1.1\%$ generalization error.
This implies that \emph{privacy violations are a threat even to well-generalized models}, since our attack is able to confidently (with at least $75\%$ confidence) identify a subset of training set members.

On the MLP models, the calibration is slightly less consistent; however, here \bayesian is able to obtain over $70\%$ precision on the models trained on the Breast Cancer Wisconsin and Hepatitis datasets.

In Figure~\ref{fig:all_apr}, we see that the recall of the uncalibrated attack is frequently over $90\%$.
When calibrating, the recall drops as precision increases, however, we believe this does not diminish the threat of the attacks because a privacy violation occurs if even a few points are confidently inferred.

While Figure~\ref{fig:rule_of_thumb} demonstrates that applying our calibration heuristic to \bayesian and \general consistently increases the precision, we see that this is not always the case for \bb.
In some cases, the precision of \bb is \emph{decreased} by increasing the decision threshold.
In fact, occasionally, the average confidence on non-members is higher than that of members, leading to a precision slightly less than $50\%$.
This may be a result of the shadow model overfitting to the hold-out data.
When we are able to increase the precision of \bb using its confidence output, the gains are less impressive, suggesting the probability outputs of \bb are less well-calibrated.

\paragraph{Performance on Well-Generalized Models}
While some of the models we used to evaluate our attacks had a generalization error of $10\%$ or more, we also evaluated on several datasets for which the learned model was far less overfit, including MNIST ($1.1\%$ generalization error), Adult ($1.2\%$), Pima Diabetes ($3.4\%$), and Breast Cancer Wisconsin ($4.3\%$).
While on PD and BCW, our attacks only slightly outperform \naive, on MNIST and Adult, our attacks do substantially better:
on the model trained on Adult, \general achieves an advantage $2.6$ times greater than the advantage achieved by \naive.
Even more impressively, on MNIST, \general and \bayesian achieves an advantage $3.5$ and $12.5$ times greater than the advantage achieved by \naive, respectively.
On the other hand, \bb fares poorly on all of these datasets except for Adult, typically achieving less than $2\%$ advantage.
Finally, we note that the \bayesian attack on the synthetic data model (Section~\ref{sec:eval_omniscient}) achieves a non-trivial $60\%$ accuracy ($20\%$ advantage), despite the fact that the model has \emph{zero} generalization error.

In addition to the cases where our attacks achieve relatively high advantage against well-generalized models, we find that when calibrated, our attacks achieve as high as 75\% precision on MNIST, and 70\% precision on Breast Cancer Wisconsin, again underscoring the threat of privacy violations for well-generalized models.

While it is clear that a greater degree of overfitting makes it easier for an adversary to mount \emph{any} attack, the relative success of our attacks over \naive on well-generalized models suggests that the white-box information is useful even when the model does not leak information through incorrect predictions on the test set.

\begin{figure*}[t!]
	\pgfplotstableread[row sep=\\,col sep=&]{
  X    & Y \\
  nodp & 60.5 \\
  do   & 65.4 \\
  e16  &  \\
  e4   &  \\
  e1   &  \\
  e25  &  \\
}\GeneralDp

\pgfplotstableread[row sep=\\,col sep=&]{
  X    & Y \\
  nodp & 61.8 \\
  do   &  \\
  e16  & 73.55 \\
  e4   &  \\
  e1   & 52.1 \\
  e25  & 55.4 \\
}\BayesianDp

\centering
\resizebox{\textwidth}{!}{
\tikzstyle{every node}=[font=\scriptsize]
\begin{tikzpicture}
  \begin{axis}[
      ybar,
      bar width=6pt,
      width=1.1\textwidth,
      height=.6\columnwidth,
      legend style={
        at={(1.0, 0.0)},
        anchor=south east,
        legend columns=-1},
      ymajorgrids,
      ytick={0.5, 0.55, .6, 0.65, .7, .75, .8, .85, .9},
      ylabel shift = -4pt,
      grid style=dotted,
      symbolic x coords={
        bcw,
        pd,
        hep,
        gc,
        adult,
        mnist,
        lfw,
        cifar10,
        cifar100},
      ytick pos=left,
      xticklabels={BCW, PD, Hep, GC, Adult, MNIST, LFW, CIFAR10, CIFAR100},
      x tick label style={anchor=north},
      enlarge x limits=.075,
      xtick={bcw, pd, hep, gc, adult, mnist, lfw, cifar10, cifar100},
      ymin=.45,ymax=.95,
      ylabel={\small\textit{accuracy}},
      legend entries={No Defense,Dropout,0.25-DP,1-DP,4-DP,16-DP},
      legend pos=north west
    ]
    \addplot[ 
        fill=bblue,
        draw=darkgray,
        error bars/.cd,
        y dir=plus,
        y explicit
      ] coordinates{
          (bcw, .523)
          (pd, .519)
          (hep, .618)
          (gc, .652)
          (adult, .516)
          (mnist, .524)
          (lfw, .636)
          (cifar10, .741)
          (cifar100, .915)
        };

    \addplot[ 
        fill=white,
        draw=darkgray,
        error bars/.cd,
        y dir=plus,
        y explicit
      ] coordinates{
          (bcw, .504)
          (pd, .516)
          (hep, .600)
          (gc, .615)
          (adult, .509)
          (mnist, .521)
          (lfw, .618)
          (cifar10, .709)
          (cifar100, .872)
        };

    \addplot[ 
        fill=grayTwo,
        draw=darkgray,
        error bars/.cd,
        y dir=plus,
        y explicit
      ] coordinates{
          (bcw, .498)
          (pd, .499)
          (hep, .494)
          (gc, .500)
          (adult, .500)
          (mnist, .501)
          (lfw, .535) 
          (cifar10, .509)
          (cifar100, .500)
        };

    \addplot[ 
        fill=grayFour,
        draw=darkgray,
        error bars/.cd,
        y dir=plus,
        y explicit
      ] coordinates{
          (bcw, .498)
          (pd, .505)
          (hep, .507)
          (gc, .502)
          (adult, .500)
          (mnist, .502)
          (lfw, .546) 
          (cifar10, .510)
          (cifar100, .499)
        };

    \addplot[ 
        fill=graySix,
        draw=darkgray,
        error bars/.cd,
        y dir=plus,
        y explicit
      ] coordinates{
          (bcw, .505)
          (pd, .510)
          (hep, .521)
          (gc, .506)
          (adult, .501)
          (mnist, .509)
          (lfw, .551) 
          (cifar10, .521)
          (cifar100, .500)
        };

    \addplot[ 
        fill=grayEight,
        draw=darkgray,
        error bars/.cd,
        y dir=plus,
        y explicit
      ] coordinates{
          (bcw, .508)
          (pd, .516)
          (hep, .579)
          (gc, .511)
          (adult, .507)
          (mnist, .514)
          (lfw, .621) 
          (cifar10, .526)
          (cifar100, .502)
        };
  \end{axis}
\end{tikzpicture}}
	\vspace{-1em}
	\caption{\label{fig:dp_attack_acc}
		Attack accuracies against models trained with either dropout or ($\epsilon,\delta$)-differential privacy for various values of $\epsilon$.
	}
\end{figure*}
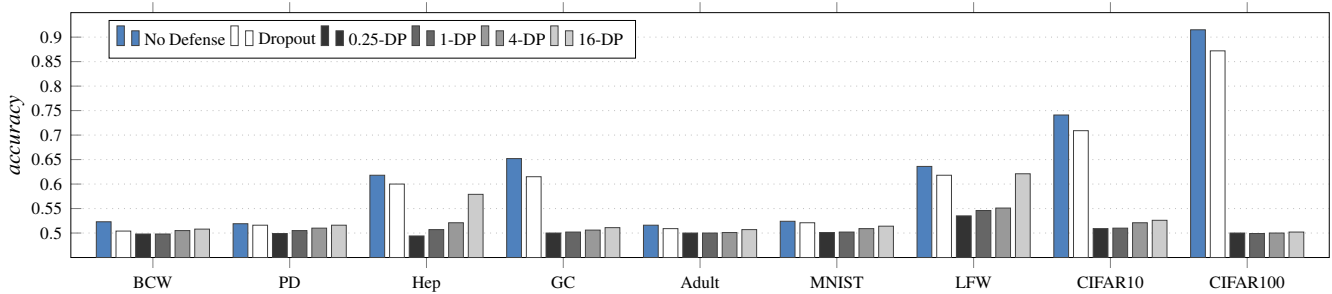

\begin{figure}[t!]

\centering
\resizebox{\columnwidth}{!}{%
\footnotesize
\begin{tabular}{llcccccc}
\toprule
\textit{dataset}       &  
& \textit{no defense} 
& \textit{dropout} 
& $\epsilon = 0.25$  
& $\epsilon = 1$   
& $\epsilon = 4$  
& $\epsilon = 16$ \\
\midrule
BCW      & \textit{train} & 0.987 & 0.982 & 0.601 & 0.654 & 0.767 & 0.778 \\
         & \textit{test}  & 0.944 & 0.961 & 0.609 & 0.675 & 0.763 & 0.808 \\
\midrule
PD       & \textit{train} & 0.789 & 0.784 & 0.680 & 0.678 & 0.681 & 0.683 \\
         & \textit{test}  & 0.756 & 0.783 & 0.673 & 0.651 & 0.649 & 0.654 \\
\midrule
Hep      & \textit{train} & 0.997 & 0.992 & 0.534 & 0.695 & 0.700 & 0.729 \\
         & \textit{test}  & 0.810 & 0.849 & 0.555 & 0.786 & 0.803 & 0.817 \\
\midrule
GC       & \textit{train} & 0.937 & 0.932 & 0.625 & 0.656 & 0.680 & 0.707 \\
         & \textit{test}  & 0.701 & 0.730 & 0.610 & 0.661 & 0.687 & 0.698 \\
\midrule
Adult    & \textit{train} & 0.861 & 0.860 & 0.501 & 0.500 & 0.500 & 0.501 \\
         & \textit{test}  & 0.849 & 0.859 & 0.500 & 0.501 & 0.500 & 0.499 \\
\midrule
MNIST    & \textit{train} & 1.000 & 0.998 & 0.107 & 0.129 & 0.243 & 0.330 \\
         & \textit{test}  & 0.973 & 0.987 & 0.106 & 0.132 & 0.251 & 0.331 \\
\midrule
LFW      & \textit{train} & 1.000 & 0.999 & 0.109 & 0.137 & 0.214 & 0.428 \\
         & \textit{test}  & 0.842 & 0.835 & 0.116 & 0.119 & 0.200 & 0.463 \\
\midrule
CIFAR10  & \textit{train} & 0.999 & 0.996 & 0.100 & 0.098 & 0.103 & 0.100 \\
         & \textit{test}  & 0.621 & 0.664 & 0.101 & 0.100 & 0.105 & 0.093 \\
\midrule
CIFAR100 & \textit{train} & 0.999 & 0.977 & 0.010 & 0.010 & 0.010 & 0.011 \\
         & \textit{test}  & 0.257 & 0.312 & 0.010 & 0.010 & 0.011 & 0.011 \\
\bottomrule
\end{tabular}}

	\caption{\label{fig:dp_tt_acc}
		Train and test accuracies for models trained with either dropout or ($\epsilon,\delta$)-differential privacy for various values of $\epsilon$.
	}
	\vspace{-1em}
\end{figure}

\paragraph{Similarity of \bb and \naive Results}
Figure~\ref{fig:all_apr} reveals that often, \bb\ has performance comparable or even worse than \naive, particularly on well-generalized target models.
This is likely a product of the attack model overfitting to idiosyncrasies in the shadow model's output that are unrelated to the target model.
On deep models with significant overfitting, \bb performs slightly better than \naive, however, we found that its behavior was not significantly different from that of \naive; for example, on LFW, \naive recovers $88\%$ of the exact correct predictions made by \bb.
This supports the intuition that the features used by the shadow model approach (i.e., the softmax outputs) are not fundamentally more well-suited to membership inference than those used by the \naive method (i.e., the correctness of the predictions).
This is perhaps unsurprising, as the softmax outputs are likely to coincide largely with the correctness of the prediction---for correct predictions, the softmax will likely have high confidence on the correct class, regardless of whether the point was a member or not; and similarly for incorrect predictions, the softmax will likely have more entropy.




\section{Defenses}\label{sec:defenses}

Concerns about privacy, underscored by concrete threats such as the attacks developed in this paper, have also motivated research to provide adequate defenses against such threats.
In this section we explore the ability of some of the commonly-proposed mitigation techniques to defend against our attack.
In particular, we focus on \emph{differential privacy}~\cite{dwork06} and regularization.
We find that, while both are useful to a degree, neither dropout nor $\epsilon$-differentially private training with a large $\epsilon$, are necessarily sufficient for mitigating the privacy risk posed by our attack.

\paragraph{Differential Privacy}
Differential privacy (DP)~\cite{dwork06} is often seen as the gold standard for private models, as models trained with differential privacy have provable guarantees against membership inference.
Namely, \citeauthor{yeom17overfitting}~\cite{yeom17overfitting} showed that, given an $\epsilon$-differentially private learning algorithm, an adversary can achieve an advantage of at most $e^\epsilon - 1$.
Differential privacy has been applied to many areas of machine learning, including logistic regression~\cite{chaudhuri09}, SVMs~\cite{rubinstein09}, and more recently, deep learning~\cite{shokri15,abadi16}.
However, current methods for ensuring differential privacy are typically costly with respect to the accuracy of the model, particularly for small values of $\epsilon$, which give a better privacy guarantee.
For this reason, in practice, $\epsilon$ is often chosen to be quite large; for example, in 2017, Apple was found to use an effective epsilon as high as $16$ in some of its routines~\cite{applePrivacy}.


We used the Tensorflow Privacy library~\cite{mcmahan18}, an implementation of the \emph{moments accountant} method~\cite{abadi16}, which guarantees $(\epsilon, \delta)$-differential privacy, to study the practical efficacy of our attack on protected models.
This method utilizes several hyperparameters from which $\epsilon$ is derived; for uniformity, we modified only the \emph{noise multiplier} to achieve the desired $\epsilon$, and used heuristics described in the original paper~\cite{abadi16} to select the remaining hyperparameters.
While a different tuning of the hyperparameters may result in a different privacy-utility trade-off, the privacy guarantee depends only on $\epsilon$ and $\delta$, \emph{not the hyperparameters directly}.
In each case, $\delta$ was selected to be smaller than $1/N$ where $N$ is the size of the dataset.

Figure~\ref{fig:dp_attack_acc} shows the effectiveness the \general attack against models trained with differential privacy for various values of $\epsilon$ on each dataset.
The train and test accuracies of the corresponding differentially-private target models are shown in Figure~\ref{fig:dp_tt_acc}.
First, we note that as expected, when $\epsilon$ decreases the adversary's effectiveness quickly declines.
However, when $\epsilon$ is large ($\epsilon = 16$), our attack occasionally performs \emph{essentially the same} on the differentially-private model as on the undefended model.
For example, on BCW, PD, and LFW, 16-DP provided less defense than simple regularization, while harming the accuracy of the model.
Similarly, on Hep, 16-DP reduced the effectiveness of \general, but not below the effectiveness of \bb on the corresponding undefended model.
These findings suggest that the practical benefits of large-$\epsilon$-differential privacy cannot be taken for granted;
in general, differential privacy may only be effective for sufficiently small $\epsilon$.

Nevertheless, it is clear that a practical adversary is unlikely to achieve performance that is tight with the theoretical bound.
For both the undefended model and the models trained with DP for $\epsilon > \ln 2 \approx 0.69$, the theoretical bound on the adversary's accuracy is $100\%$, which no attack was able to achieve.
On the other hand, for $\epsilon = 0.25$, the theoretical maximum accuracy of the adversary is $64.2\%$.
In most such cases, our attack fared far poorer than this, coming closest on LFW, where our attack achieved $53.5\%$ accuracy ($25\%$ of the theoretical maximum advantage) on the $0.25$-DP model.
Thus, we conclude that because the accuracy of a real adversary is not likely to be tight with the worst-case guarantee, it is indeed pragmatic to select a somewhat large $\epsilon$.
However, our evaluation shows that $\epsilon$ should not be chosen to be too large, or else the operative benefits of differential privacy may be lost.
Furthermore, the success of a given value of $\epsilon$ appears to vary across different datasets and models.
One must therefore be careful when making a practical selection for $\epsilon$;
to this end, we suggest that our attack may be useful in assessing which values of $\epsilon$ are appropriate for a given application.



An apparent drawback of the examined method for obtaining differential privacy, revealed in our evaluation, is the steep cost in performance (Figure~\ref{fig:dp_tt_acc}), which is particularly high for small $\epsilon$.
Despite the fact that our attack became far less effective for small $\epsilon$, this cost limits the practicality of the defense, highlighting the need for more research in this area.
The results we find here align with recent work~\cite{jayaraman19dpInPractice}, in which \citeauthor{jayaraman19dpInPractice} showed that the privacy leakage tends to increase as $\epsilon$ becomes large enough to avoid a significant loss in accuracy.
Indeed, only on the German Credit dataset did 16-DP provide a good defense while nearly maintaining the accuracy of the unprotected model.
In the other cases we evaluated, either our attack performed comparably on the DP and unprotected models, or the accuracy of the private model was significantly lower than that of the unprotected model.

\citeauthor{abadi16}~\cite{abadi16} mitigate the high cost in accuracy by first pre-training on public data, and then fine-tuning only the top layers with differential privacy on the private training set.
While this public transfer learning approach may not always be possible, it has two key benefits, the first being that the resulting model's performance is far less poor.
Second, only the final layers of such a model are trained on the private data, and thus our attack may only be able to effectively target those layers.
Our experiments in Section~\ref{sec:eval_layers} show that our attack is far more effective when all layers are leveraged, and that the earlier layers often account for a sizable portion of the information leakage.
This suggests that, when possible, a transfer learning scheme like that of \citeauthor{abadi16} could be a practical defense.


\paragraph{Regularization}
Given the connection between membership inference and overfitting, regularization, such as dropout~\cite{srivastava14}, which aims to reduce overfitting, has also been proposed to combat membership inference.
Generalization alone is not sufficient to protect against membership inference~\cite{yeom17overfitting}, and in fact, our empirical results (Section~\ref{sec:eval}) show that we can successfully attack even models with negligible generalization error; however, dropout has been shown not only to reduce overfitting, but to strengthen privacy guarantees in neural networks~\cite{jain15}.
Figure~\ref{fig:dp_attack_acc} shows the accuracy of our attack with and without dropout.
We find that dropout does not significantly impact the accuracy of our attack in most cases.
However, as opposed to DP, dropout is typically \emph{beneficial} to the performance of the model, while providing a modest defense.
In this light, regularization (including dropout) may in fact be the more practical defensive measure, insofar as it improves test accuracy, because better generalization does appear to make membership more difficult, though clearly not impossible, for an attacker.

Still, we warn that this may not be universally true of all forms of regularization, even regularization that improves generalization---as we have demonstrated, a model can still leak membership information through its parameters while making correct predictions on unseen points.


\paragraph{Defenses in the Black-box Setting}
For membership inference in the black-box setting, \citeauthor{shokri16shadow}~\cite{shokri16shadow} also propose a number of other possible defenses, such as restricting the prediction vector to the top $k$ classes, or increasing the entropy of the prediction vector via increasing the normalization temperature of the softmax.
However, these defenses are easily circumvented in the white-box setting, as the pre-modified outputs are still available to an attacker in this threat model.

Similarly, \citeauthor{ndss19salem}~\cite{ndss19salem} propose a defense called \emph{model stacking}, in which two models are trained separately on the training data and a third model makes predictions based on the outputs of the first two.
While \citeauthor{ndss19salem} found this to be an effective defense against black-box approaches, this defense is likewise circumvented in the white-box setting, as the initial two models are available to the attacker.


\section{Related Work}\label{sec:related}\label{sec:related_work}

There is extensive prior literature on privacy attacks on statistical summaries.
\citeauthor{homer08resolving}~\cite{homer08resolving} proposed what is considered the first membership inference attack on genomic data in 2008.
Following the work by \citeauthor{homer08resolving}, a number of studies~\cite{wang-snp,sankararaman2009genomic,ElEmam11,Gymrek321,shringarpure15} have looked into membership attacks on statistics commonly published in genome-wide association studies.
In a similar vein, \citeauthor{Komarova}~\cite{Komarova} looked into partial disclosure scenarios, where an adversary is given fixed statistical estimates from combined public and private sources and attempts to infer the sensitive feature of an individual referenced in those sources.

More recently, membership inference attacks have been applied to machine learning models.
\citeauthor{AtenieseFMSVV13}~\cite{AtenieseFMSVV13} demonstrated that given access to the parameters of support vector machines (SVMs) or Hidden Markov Models (HMMs), an adversary can extract information about the training data.

As deep learning has become more ubiquitous, membership inference attacks have been particularly directed at deep neural networks.
A number of different recent works~\cite{shokri16shadow,ndss19salem,yeom17overfitting,long17,long18generalized,nasr18whitebox} have taken different approaches to membership inference against deep networks in a standard supervised learning setting.
Additionally, \citeauthor{hayes17logan}~\cite{hayes17logan} have studied membership inference against generative adversarial networks (GANs); and others~\cite{nasr18whitebox,hitaj17,melis18collaborative} have studied membership inference in the context of collaborative, or federated, learning.

\paragraph{Black-box attacks}
We study membership inference as it applies to deep networks in classic supervised learning problems.
Most of the prior work in this area~\cite{shokri16shadow,ndss19salem,yeom17overfitting,long17,long18generalized} has used the \emph{black-box} threat model.
\citeauthor{yeom17overfitting}~\cite{yeom17overfitting} showed that generalization error necessarily leads to membership vulnerability; a natural consequence of this is that a simple ``naive'' attack (\naive), which predicts a point is a member if and only if it was classified correctly, can be found to be quite effective on models that overfit to a large degree.
Other approaches have leveraged not only the predictions of the model, but the confidence outputs.
A particularly canonical approach, along these lines, is the attack introduced by \citeauthor{shokri16shadow}~\cite{shokri16shadow} (\bb).
In this approach, a \emph{shadow model} is trained on half of $\tilde S$, $\tilde S_{in}$, and an \emph{attack model} is trained on the the outputs of the shadow model on its training data, $\tilde S_{in}$ (labeled $1$), and the remaining data $\tilde S\setminus\tilde S_{in}$ (labeled $0$).
Shadow models leverage the disparity in prediction confidences on training instances the target model has overfit to, and have been shown to be successful at membership inference on models that have sufficiently high generalization error.
A few other membership inference approaches~\cite{ndss19salem,hayes17logan} have made use of this same technique.

Despite the fact that shadow model attacks leverage more information than the naive attack, we find in our evaluation (Section~\ref{sec:eval}) that often, the shadow model attack fails to outperform the naive attack.
One potential reason for this finding is that the learned attack model used by this approach to distinguish between the shadow model's outputs on members and non-members may be itself subject to overfitting.
This may be especially true if the attack model picks up on behavior particular to one of the shadow models rather than the true target model.
Furthermore, the confidence and entropy of the target model's softmax output is likely to be closely related to whether the target model's prediction was correct or not, meaning that the softmax outputs may not provide substantially different information from that used by \naive.





\paragraph{White-box attacks}
In some settings, it may be realistic for an attacker to have white-box access to the target model.
Intuitively, while some information is leaked via the behavior of a model, the details of the structure and the parameters of the model are clear culprits for information leakage.
Few prior approaches have successfully leveraged this extra information.
While \citeauthor{hayes17logan}~\cite{hayes17logan} describe a white-box attack in their work on membership inference attacks applied to GANs, the attack uses access only the outputs of the discriminator portion of the GAN, rather than the learned weights of either the discriminator or the generator; thus their approach is not white-box in the same sense.
Meanwhile, \citeauthor{nasr18whitebox}~\cite{nasr18whitebox} demonstrated that a simple extension of the black-box shadow model approach to utilize internal activations does not result in higher membership inference accuracies than the original black-box approach.
This is perhaps unsurprising, as the internal units of the shadow models are not likely to have any relation to those of the target model (see Section~\ref{sec:deep}).

Recently, \citeauthor{nasr18whitebox}~\cite{nasr18whitebox} provided a white-box attack that leverages the gradients of the target model's loss function with respect to its weights, which SGD approximately brings to zero on the training points at convergence.
In contrast to our work, \citeauthor{nasr18whitebox} use a further relaxed threat model, in which the attacker has access to as much as \emph{half of the target model's training data}.
We suggest an approach that is quite different from that of \citeauthor{nasr18whitebox}.
Our approach does not require this extra knowledge for the attacker, and thus falls under a more restrictive threat model, in which, to our knowledge, no other effective white-box attacks have been proposed.

\section{Conclusions and Future Work}\label{sec:future}

Our work is the first to fully leverage white-box information to improve membership inference attacks against deep networks (in the standard threat model where the adversary is assumed not to have any examples of true training points).
In particular, \emph{our analysis sheds light on a fundamental mechanism of overfitting that can be leveraged by an adversary to compromise a model's privacy in a concrete way}.
We use this analysis of how feature usage can lead to information leakage to construct a new white-box attack, which our evaluation demonstrates improves upon the previous state-of-the-art, particularly because it can be reliably calibrated for high precision, 
even on some well-generalized models.

Subsequently, we used our attack to evaluate commonly-proposed privacy defenses.
Perhaps most interestingly, experiments utilizing our attack reveal a nuanced story regarding differential privacy.
When setting $\epsilon$ to small values, the attack was successfully mitigated but the utility of the resulting model quickly diminished; while when $\epsilon$ was increased sufficiently to mitigate the loss in utility, the attack sometimes achieved close to the same accuracy as on the undefended model.
This suggests that there is still considerable work to be done in developing effective defenses against privacy attacks---we anticipate that the insights gained from our approach will contribute to designing such defenses.


\paragraph{Acknowledgment} 
This material is based on work supported by the National Science Foundation under Grants No. CNS-1704845 and CNS-1801391.

\bibliographystyle{plainnat}
{\small \bibliography{bibliography}}

\appendix

\section{Proof of Theorem~\ref{thm:opt_wb}}\label{app:proof}
\noindent
We begin with the expression for $m^y(x)$ and apply Bayes' rule to obtain Equation~\ref{eq:bayes_rule}.
\begin{align}
    m^y(x) 
    &= \p{T\ |\ X = x, Y = y} = \frac{\p{X=x\ |\ T, Y=y}\p{T}}{\p{X=x\ |\ Y=y}} \label{eq:bayes_rule} 
\end{align}
Next, we express Equation~\ref{eq:bayes_rule} as a logistic (or, sigmoid) function, $\sigmoidf(x) := (1 + e^x)^{-1}$.
We assume that $\p{T} = \half$, and thus $\p{X=x\ |\ Y=y}$ can be written as $\frac{1}{2}\left(\p{X=x\ |\ T, Y=y} + \p{X=x\ |\ \neg T, Y=y}\right)$, by the law of total probability.
We then divide by the numerator in Equation~\ref{eq:bayes_rule}, yielding an expression that can be written as a logistic function (\ref{eq:to_sigmoid}) by noting that for $x > 0$, $\exp(\log\ x) = x$. 
\begin{align}
    (\ref{eq:bayes_rule})
    &= \frac{\p{X=x\ |\ T, Y=y}}{\left(\p{X=x\ |\ T, Y=y} + \p{X=x\ |\ \neg T, Y=y}\right)} \nonumber\\[4pt]
     &= \left(1 + \frac
        {\p{X=x | \neg T, Y=y}}
        {\p{X=x | T, Y=y}}\right)^{-1} \nonumber\\[4pt]
    &= \left(1 + \exp\left(\log\frac
        {\p{X=x | \neg T, Y=y}}
        {\p{X=x | T, Y=y}}\right)\right)^{-1} \nonumber\\[4pt]
    &= \sigmoid{\log\frac{\p{X=x\ |\ T, Y=y}}{\p{X=x\ |\ \neg T, Y=y}}} \label{eq:to_sigmoid}
\end{align}
We notice that $\p{X=x\ |\ T, Y=y}$ is the probability of having drawn $x$ from $\hat\distrib$, given class, $y$, and similarly, $\p{X=x\ |\ \neg T, Y=y}$ is the probability of having drawn $x$ from $\distrib^*$, given class, $y$.
Using the Naive-Bayes assumption, i.e., that conditioned on the class, $y$, the individual features, $x_j$, are independent, we obtain Equation~\ref{eq:use_nb_assumption}.
\begin{align}
    (\ref{eq:to_sigmoid})
    &= \sigmoid{\log\prod_j{\frac
        {\N(x_j\ |\ \hat\mu_{yj}, \hat\sigma_j^2)}
        {\N(x_j\ |\ \mu^*_{yj}, \sigma_j^{*2})}}} \label{eq:use_nb_assumption}
\end{align}
We then re-write the $\log$ of the product as a sum over the $\log$, and observe that the sum can be written as a dot product as in Equation~\ref{eq:mi_model}, which gives the parameters of the Bayes-optimal model for $m^y(x)$.
\begin{align}
    (\ref{eq:use_nb_assumption})
    &= \sigmoid{\sum_j{
        \frac{(x_j - \mu_{yj}^*)^2}{2\sigma_j^{*2}} - \frac{(x_j - \hat\mu_{yj})^2}{2\hat\sigma_j^2} +
        \log\left(\frac{\sigma_j^{*}}{\hat\sigma_j}\right)}} \nonumber \\[4pt]
    &= \sigmoid{v^{y\ T}x^2 + w^{y\ T}x + b^y} \label{eq:mi_model} \\
    \text{where} \nonumber \\
    v^y_j &= \frac{1}{2\sigma_j^{*2}}-\frac{1}{2\hat\sigma_j^2}
    \qquad 
    w^y_j = \frac{\hat\mu_{yj}}{\hat\sigma_j^2}-\frac{\mu_{yj}^*}{\sigma_j^{*2}} \nonumber \\
    b^y &= \sum_j{
        \left(\frac{\mu_{yj}^{*2}}{2\sigma_j^{*2}}-\frac{\hat\mu_{yj}^2}{2\hat\sigma_j^2}\right)} +
        \log\left(\frac{\sigma_j^{*}}{\hat\sigma_j}\right) \nonumber
\end{align}
Finally, by assumption the variance is the same in $S$ as in the general distribution, i.e., $\hat\sigma_j = \sigma^*_j = \sigma_j$, for all features, $j$. Thus, $v^y$ from Equation~\ref{eq:mi_model} becomes zero, so we are left with a linear model for $m^y$, with weights, $w^y$, and bias, $b^y$, given by Equation~\ref{eq:bayes-predictor}.
$\Box$





\end{document}